\pdfoutput=1

\documentclass[11pt]{article}

\usepackage{ACL2023}
\usepackage{times}
\usepackage{latexsym}

\usepackage[T1]{fontenc}

\usepackage[utf8]{inputenc}

\usepackage{microtype}

\usepackage{inconsolata}

\usepackage{graphicx}
\usepackage{pifont}
\usepackage[linesnumbered,ruled,vlined]{algorithm2e}
\usepackage{algpseudocode}

\usepackage{amsmath, bm}
\usepackage{amsfonts}
\usepackage{enumitem}
\usepackage{float} 
\newtheorem{myDef}{Definition}

\usepackage{listings}
\usepackage{color}

\definecolor{dkgreen}{rgb}{0,0.6,0}
\definecolor{gray}{rgb}{0.5,0.5,0.5}
\definecolor{mauve}{rgb}{0.58,0,0.82}

\title{Learning to Substitute Spans towards Improving Compositional Generalization}

\author{Zhaoyi Li$^{1,2}$,  Ying Wei$^{3}$\thanks{\ \ Corresponding authors} \and 
Defu Lian$^{1,2}$\footnotemark[1] \\
$^{1}$School of Computer Science and Technology, University of Science and Technology of China\\
$^{2}$State Key Laboratory of Cognitive Intelligence, Hefei, Anhui, China\\
$^{3}$Department of Computer Science, City University of Hong Kong\\
\texttt{lizhaoyi777@mail.ustc.edu.cn, yingwei@cityu.edu.hk, liandefu@ustc.edu.cn}}
\begin{document}
\maketitle

\begin{abstract}
Despite the rising prevalence of neural sequence models, 
recent empirical evidences 
suggest their deficiency in compositional generalization. 
One of the current de-facto solutions to this problem is compositional data augmentation, aiming to incur additional compositional inductive bias.
Nonetheless, the improvement offered by existing handcrafted augmentation strategies is limited when successful systematic generalization of neural sequence models requires multi-grained compositional bias (i.e., not limited to either lexical or structural biases only) or 
differentiation of training sequences in an imbalanced difficulty distribution.
To address the two challenges, we first propose a novel compositional augmentation strategy dubbed \textbf{Span} \textbf{Sub}stitution (SpanSub) that enables multi-grained composition of substantial substructures 
 in the whole training set.
Over and above that, we introduce the \textbf{L}earning \textbf{to} \textbf{S}ubstitute \textbf{S}pan (L2S2) framework which empowers the learning of span substitution probabilities in SpanSub 
in an end-to-end manner by maximizing the loss of neural sequence models, so as to outweigh those challenging compositions with elusive concepts and novel surroundings.
Our empirical results on three standard compositional generalization benchmarks, including SCAN, COGS and GeoQuery (with an improvement of at most 66.5\%, 10.3\%, 1.2\%, respectively),
demonstrate the superiority of  
SpanSub, 
L2S2 and their combination.
\end{abstract}
\section{Introduction}
\label{sec:intro}
\begin{figure}[h]
\centering 
\includegraphics[width=0.47\textwidth]{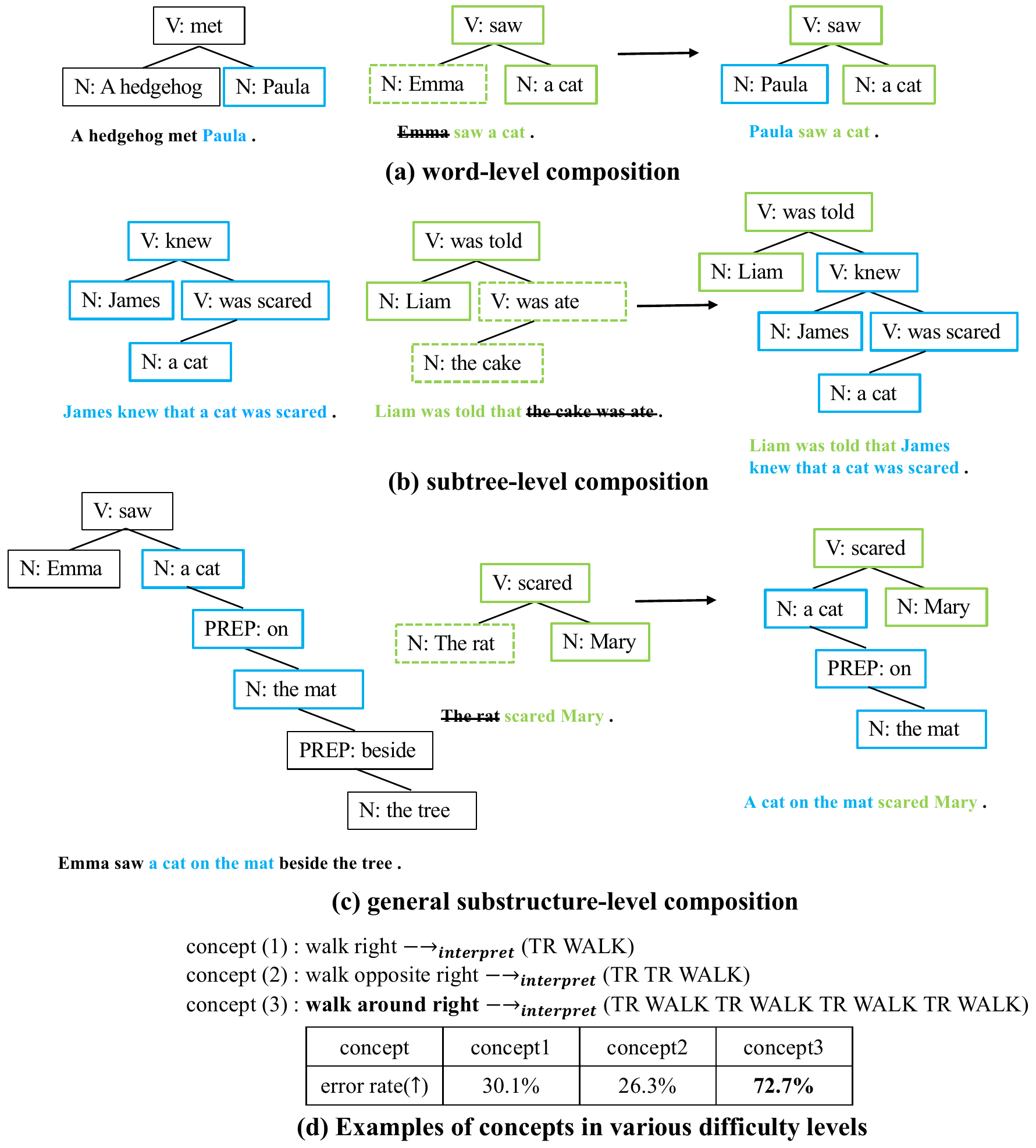}
\caption{(a), (b) and (c) illustrate three distinct compositional generalization types in COGS~\cite{cogs}, which require word-level, subtree-level and general substructure-level recombinations of training data, respectively. Besides, (d) shows concepts in distinct difficulty in the SCAN~\cite{scan} dataset, where the interpretation of \emph{walk around right} is much more complex than that of the other two concepts.}
\label{figure:intro}
\vspace{-0.1in}
\end{figure}

The secret for human beings to learning so quickly with little supervision has been
demonstrated to be associated with the powerful ability of \emph{systematic generalization}, being capable of producing an infinite number of novel combinations on the basis of known components~\cite{chomsky}.
In stark contrast, a large body of recent evidence suggests that current state-of-the-art neural sequence models lack of adequate power for compositional generalization (\emph{a.k.a.,} systematic generalization)~\cite{scan, pretrain.vs.special}. 
For instance, a model which has observed the two training sentences of ``\emph{look opposite right} twice and jump right thrice'' and ``\emph{walk around right} and run twice'' likely fails to understand the testing sentence of ``\emph{walk around right} twice and jump right thrice''.
Sharpening the compositional generalization ability of 
neural sequence models is beyond important to fill the gap with human-like natural language understanding, catalyzing not only better performances but also fewer expensive annotations.

Inspired by the tight relationship between compositionality and group-equivariance of neural models~\cite{permutation-eq,lexsym,Basu2022EquiTuningGE}, a series of compositional data augmentation solutions have made great strides via injecting compositional inductive bias into neural sequence models~\cite{geca, seqmix, lexsym, subs, mutual}. 
The key idea behind compositional data augmentation is to 
substitute a part in one original training example with a part from another training example, thus composing a novel example that complements the training data with compositional bias. 
Introducing comprehensive enough comositional bias to embrace a diversity of testing tasks, however, is not trivial. 
First, the ``part''\footnote{We use the words of ``part'', ``concept'', and ``span'' later interchangeably.} to be substituted out and in is expected to be in multiple levels, ranging from words~\cite{lexsym} in Fig.~\ref{figure:intro}(a), to complete substrees~\cite{subs} in Fig.~\ref{figure:intro}(b), to more general substructures in Fig.~\ref{figure:intro}(c). How to develop an augmentation method that flexibly accommodates multiple levels of parts remains an open question.
Second, the ``parts'' are uneven in their difficulty levels. As
shown in Fig.~\ref{figure:intro}(d), though the numbers of both training and testing sentences containing the three concepts in the SCAN MCD split are comparable and we have applied compositional data augmentation via the proposed SpanSub (which will be detailed later), the predicted error rates of testing sentences grouped by the three concepts still differ significantly, which is in alignment with the observations in ~\cite{bogin-etal-2022-unobserved}. There is an urgent need to augment with difficulty awareness and allow more compositions on the challenging concepts (e.g., concept 3 in Fig.~\ref{figure:intro}(d)).


To conquer the two challenges, we first propose a novel compositional data augmentation scheme 
SpanSub that substitutes a \emph{span} in a training sentence with one in another sentence, where a span refers to a consecutive fragment of tokens that subsumes all multi-grained possibilities of a word, a subtree, as well as a more general substructure. The core of SpanSub lies in extraction of such spans and identification of exchangeable spans, towards which we define the exchangeability of spans by the exchageability or syntactic equivalence of their first and last tokens.
On top of this, we propose the L2S2 framework made up of
a L2S2 augmenter, which is a differentiable version of SpanSub with all substitution actions equipped with probabilities. By training down-stream neural sequence models to evaluate the difficulty of various spans and maximizing their losses, the L2S2 framework seeks to train the L2S2 augmenter to tip the scales of those substitution actions contributing challenging compositions by elusive spans and novel surroundings.


In summary, the main contributions of this paper are three-fold.
\begin{itemize}[leftmargin=*,noitemsep,topsep=0pt]
    \item SpanSub is the first to explore span-based compositional data augmentation, thus flexibly supporting multi-grained compositional bias;
    \item L2S2 as a differentiable augmentation framework first empowers difficulty-aware composition, being compatible with various down-stream models.
    \item We have empirically demonstrated the superiority of SpanSub, L2S2, and their combination on three standard benchmarks (SCAN, COGS and GeoQuery) with improvements of at most $66.5\%$, $10.3\%$ and $1.2\%$ over prior part, respectively.\footnote{Code available at \url{https://github.com/Joeylee-rio/Compgen_l2s2}}
\end{itemize}
\section{Related Work}

\textbf{Compositional generalization in neural sequence models}
A large body of literature pursues various ways of introducing compositional inductive bias into neural sequence models, in a bid to improve systematic generalization. 
The first category of studies, e.g., CGPS~\cite{primsub}, SyntAtt~\cite{synt_att}, GroupEqu~\cite{permutation-eq}, customizes neural architectures that promote lexical generalization via explicit disentanglement of  the meaning of tokens.
The second strand aims to align words or substructures in the input sequences with their counterparts in the output sequences by 
auxiliary tasks 
(e.g., 
IR-Transformer~\cite{IR-transf}), 
additional architectural modules 
(e.g., LexLearn~\cite{lexlearn}),
as well as extra objectives imposed on attention layers (e.g., SpanAtt~\cite{span_attention}).
Third, the works of Meta-seq2seq~\cite{meta-seq2seq}, Comp-MAML~\cite{meta-comp}, and MET~\cite{mutual} resorts to the meta-learning paradigm 
to directly encourage compositional generalization of neural models. 
Last but not least,
compositional data augmentation that composes in-distribution data to accommodate out-of-distribution compositional sequences has been empirically demonstrated to enjoy not only the performance but also the model-agnostic benefits.
The explored principles for augmentation include exchangeability of tokens in the same context (e.g., GECA~\cite{geca}),  
token-level mixup~\cite{mixup} (e.g.,
SeqMix~\cite{seqmix}), 
group-equivariance of 
language models~\cite{Basu2022EquiTuningGE} 
by substituting training tokens (e.g., LexSym~\cite{lexsym}, Prim2PrimX~\cite{mutual}) or subtrees (e.g., SUBS~\cite{subs}) 
with virtual or off-the-shelf tokens or substrees.
Note that the aforementioned approaches guarantee the validity of composed sequences by following the widely accepted alignment practices in NLP, e.g., SpanTree~\cite{spanparse} and FastAlign~\cite{fastalign}.
Our work further pushes ahead with compositional data augmentation by (1) substituting
spans, which offers more diverse and flexible generalization than 
substituting monotonous tokens or subtrees, and (2) endowing the augmentation strategy to be differentiable and learnable in an end-to-end manner, which dynamically adapts to the difficulty of
down-stream neural sequence tasks. 
\section{Span Substitution}
\label{sec:spansub}
\begin{figure}
\centering 
\includegraphics[width=0.47\textwidth]{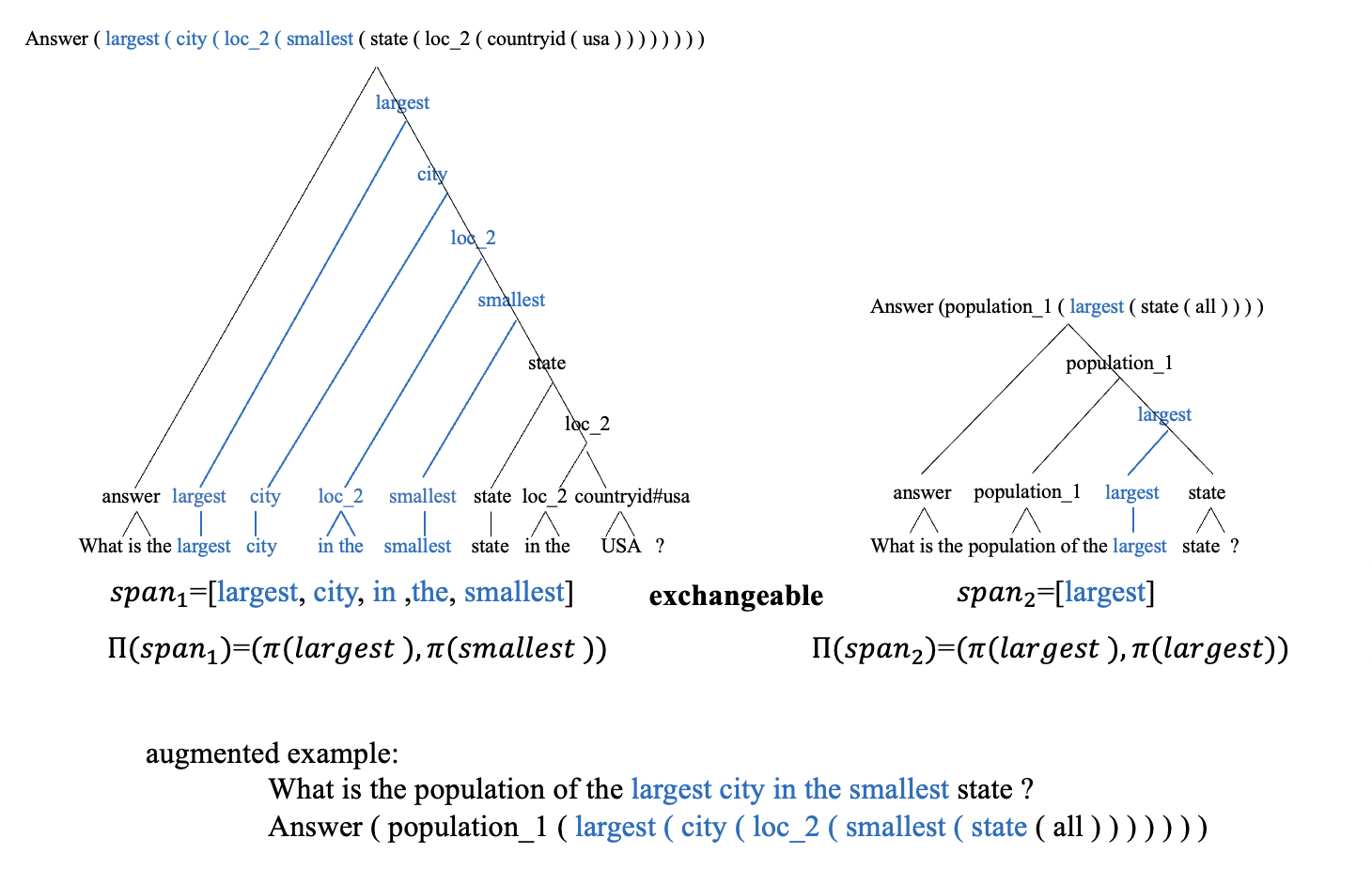}
\caption{An augmentation example by SpanSub. 
SpanSub substitutes a span ``largest'' with another span ``largest city in the smallest'', and augments a new question ``What is the population of the largest city in the smallest state?''.}
\label{figure:spansub}
\vspace{-0.1in}
\end{figure}
We propose SpanSub to generate novel examples through exchanging multi-grained spans, which refer to consecutive fragments in input sequences, of the same equivalence class between training examples as shown in Fig.~\ref{figure:spansub}. 
Before proceeding to the details of SpanSub, we first introduce two preprocessing prerequisites for SpanSub, including extraction of span alignment and inference of the equivalence class of a word. 
On top of these, we present our substitution strategy that dictates the equivalence and exchangeability between spans.
\subsection{Preprocessing}
\label{sec:preprocess}
The techniques of 
extracting span alignment from paired linguistic data and identifying syntactically equivalent words 
(e.g., Part-of-Speech tagging) have been well studied in the NLP community.
Following the practice in
a wealth of literature on compositional augmentation~\cite{lexsym, subs, mutual}, 
we also directly adapt the off-the-shelf techniques, which we introduce as below for self-contained purpose, to preprocess rather than delving into them. 
More details and results of preprocessing for all the datasets are available in Appendix~\ref{append:data_preprocess}. \\
\textbf{Extraction of span alignment}
Span alignment refers to establish the correspondence between spans in the input sequence (e.g.,  ``largest city in the smallest'') and their counterparts (e.g.,   ``largest(city(loc\_2(smallest())))'') in the output sequence of a training example. 
For the SCAN dataset, we extract span alignment by extending SimpleAlign~\cite{lexlearn} that targets single words (e.g., \emph{jump $\rightarrow$ JUMP} \emph{right $\rightarrow$ TURN\_RIGHT}) to support alignment of consecutive fragments (e.g., \emph{jump right $\rightarrow$ TURN\_RIGHT JUMP}). 
As there always exists a deterministic function program~\cite{IR-transf,subs} that transforms the output sequence $y$ to a tree for COGS and GeoQuery, we resort to the intermediate representation~\cite{IR} of COGS from~\cite{IR-transf} and the span tree of GeoQuery from~\cite{spanparse} to map the input sequence $x$ to the tree form $T$, respectively.
The tree $T$, in such a way, serves as a bridge to align the input and output.\\
\textbf{Inference  of the equivalence class of a word}
\label{preprocess:cluster_token}
The aim is to infer the equivalence class of a word $w$, i.e., $\pi(w)$, according to the cluster it belongs to. Exemplar clusters include verbs and nouns. Fortunately, the COGS dataset has intrinsic clusters of words by their 
tree structure representations.
As for SCAN and GeoQuery, we follow~\cite{lexsym, mutual} to assign those words sharing the context into a single cluster. 
For example, the words of ``largest'' and ``smallest'' fall into the same cluster in Fig.~\ref{figure:spansub}.
\subsection{Substitution Strategy}
\begin{figure}
\centering 
\includegraphics[width=0.47\textwidth]{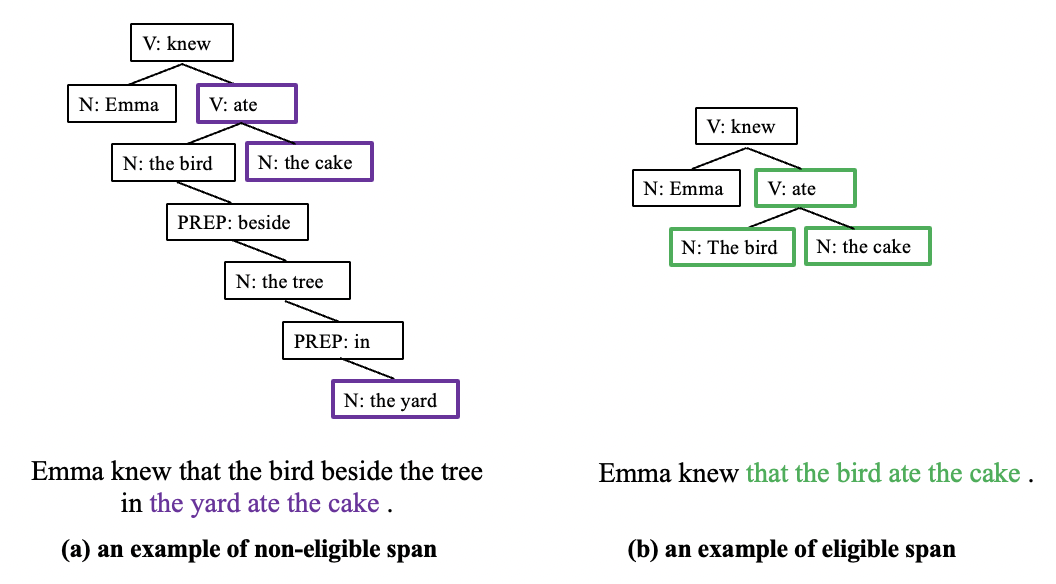}
\caption{Examples of non-eligible and eligible spans in COGS. (a) shows a non-eligible span which corresponds to an union set of disconnected fragments of the tree.}
\label{figure:eligible}
\vspace{-0.1in}
\end{figure}
The equivalence or exchangeability of spans, which a substitution strategy aims to establish, boils down to answering the following two questions: (1) what is an eligible span? (2) how to define the equivalence? First, given a consecutive span ${s = [w_p, w_{p+1}, ..., w_{p+k}]}$ where ${w_{p+i}} \ (0\leq i\leq k)$ represents a semantic unit (i.e., a word with semantic meaning), we define the span to be eligible if and only if it is semantically self-contained and unitary. Fig.~\ref{figure:eligible} shows a non-eligible span example ``the yard ate the cake'' which corresponds to an union set of two disconnected fragments of the tree and has an ambiguity (the subject of ``ate'' should be ``the bird'' rather than ``the yard''.). Such constraints imposed on eligible spans prevent substitutions with duplicate or missing parts. Due to page limit, we leave the formal mathematical definition of an eligible span into Appendix~\ref{append:define_spansub}. 

Second, we formalize a heuristic rule to define the equivalence class of an eligible span $s$ as the combined equivalence classes of its first and last token, i.e.,

\vspace{-0.15in}
\small
\begin{align}
\Pi(s) \!=\! \Pi([w_p, w_{p+1}, ..., w_{p+k}]) \!=\! (\pi(w_p),\pi(w_{p+k})), 
\end{align}
\normalsize
where $\pi$ indicates the equivalence class of a single word as defined in Section \ref{preprocess:cluster_token}. 
By defining as above, it is legal to substitute a span ${s_1}$ with another span ${s_2}$ if and only if (1) both $s_1$ and $s_2$ are eligible according Definition 1 in Appendix~\ref{append:define_spansub} and (2) $\Pi(s_1)=\Pi(s_2)$. Detailed pseudo codes of SpanSub is also available (i.e., Alg.~\ref{algo:spansub}) in Appendix~\ref{append:define_spansub}.

When dealing with tree structured tasks like GeoQuery and COGS, there are two special cases that need to be considered:
\begin{itemize}[leftmargin=*,noitemsep,topsep=0pt]
    \item ${s}\!=\![w_p]$ (e.g., ``largest'' in Fig.~\ref{figure:spansub}) degenerates to a single word: we specify that ${s}$ can only be substituted with another span $s'$ (either degenerated or undegenerated) with $\Pi(s')=[\pi(w_p),\pi(w_p)]$. 
    \item $s$ is a subtree with its root token $w_r$: we specify that ${s}$ can exchange with either another subtree $s'$ with $\Pi(s')=[\pi(w_r),\pi(w_{r})]$ or another span $s'$ with $\Pi(s')=[\pi(w_p),\pi(w_{p+k})]$).
\end{itemize}
\section{Learning to Substitute Spans (L2S2)}
\begin{figure}
\centering 
\includegraphics[width=0.47\textwidth]{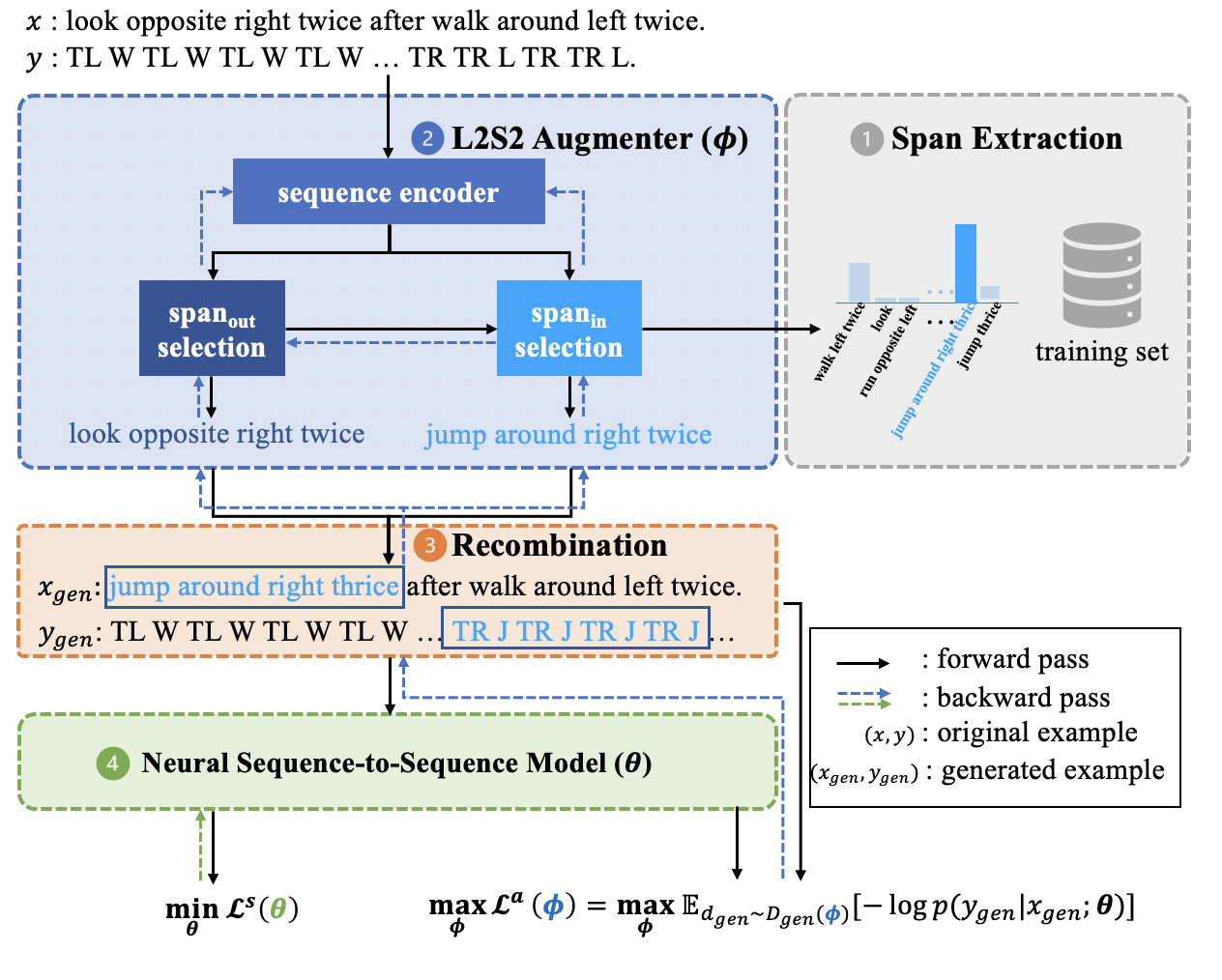}
\caption{Illustration of L2S2 framework.}
\label{figure:l2s2}
\end{figure}

Beyond the benefit of multi-grained compositional bias introduced by SpanSub, the following three observations lead us to take a step further towards augmentation with attention on challenging spans.
(1) The 
distinct combinations for a linear number of distinct spans could be as many as the super-linear number~\cite{find_needles}. 
(2) The spans constitute both easy-to-comprehend and elusive ones, while oftentimes elusive ones are so rare that those combinations by them account for a very small portion.
(3) It is imperative to increase the percentage of these minority combinations to improve
the compositional generalization in a broad range of down-stream tasks.
Concretely,
we 
introduce an online and optimizable L2S2 framework consisting of a L2S2 augmenter that 
inherits the idea of span substitution with SpanSub. 
More importantly, 
through maximizing the loss of down-stream neural sequence models, we learn span substitution probabilities in the upstreaming L2S2 augmenter to put high values on those chanllenging compositions of elusive spans and novel surroundings. The overview of the L2S2 framework is shown in Fig.~\ref{figure:l2s2}. 

\subsection{Parameterizing the L2S2 Augmenter}
\label{section:parameterization}

Given a training example $\bm{d}\!=\!{(x, y)}$, the objective of the L2S2 augmenter is to synthesize 
a new example ${\bm{d}_{gen}}\!=\!({x_{gen}}, {y_{gen}})$ via a sequence of~two actions ${\bm{a}\!=\!(a_{out},a_{in})}$: (1) $a_{out}$ which selects the
span ${s_{out}}$ to be swaped out from the span set $\mathcal{S}_1\!=\!\{s_1^i\}_{i=1}^u$ extracted from ${x}$\footnote{We can also identify spans in the ${y}$. This depends on the task type.}, and (2) 
$a_{in}$ which selects the span ${s_{in}}$ to be swapped in from the span set $\mathcal{S}_2\!=\!\{s_2^i\}_{i=1}^v$ extracted from the whole training dataset, following $a_{out}$.
Note that the preprocessing and span set extraction procedures are similar with Section~\ref{sec:spansub}, and $\mathcal{S}_1\!\subset\!\mathcal{S}_2$.
Once ${s_{out}}$ and ${s_{in}}$ are~selected, we have ${\bm{d}_{gen}}$ 
via recombination, i.e.,
\begin{itemize}[itemsep=0pt]
    \item ${x_{gen}}$ = ${x}$.replace(${s_{out}}$,${s_{in}}$),
    \item ${y_{gen}}$ = ${y}$.replace(${align(s_{out})}$,${align(s_{in})}$),
\end{itemize}
where replace($p,q$) denotes $p$ is replaced with $q$.

The probability of generating an ideal ${d_{gen}}$ based on ${d}$ 
is intuitively 
factorized as follows:
\begin{align}
  &p({\bm{d}_{gen}}|\bm{d};\bm{\phi})=p(\bm{a}|\bm{d};\bm{\phi})=p({(a_{out},a_{in})}|\bm{d};\bm{\phi}) \nonumber \\
 &=p({a_{out}}|\bm{d};\bm{\phi})\cdot p({a_{in}}|{a_{out}},\bm{d};\bm{\phi}) 
\end{align}
where $\bm{\phi}$ denotes the parameters of the L2S2 augmenter. In the following, we will detail how to model the two probabilities, during which we will introduce the 
the three parts that constitute $\bm{\phi}$.\\
\textbf{Parameterizing $p({a_{out}}|\bm{d};\bm{\phi})$ for selection of spans to be substituted out} 
Whether a span should be swapped out conditions on the equivalence class and the surroundings of the span, which are dictated by the representation of the span and that of the original training sequence ${x}$, respectively. To this end, we formulate the probability distribution $p({a_{out}}|\bm{d};\bm{\phi})$ over all $u$ candidate spans in $S_1$ as follows,
\begin{align}
\label{eq:out_distribution}
p({a_{out}}|\bm{d};\bm{\phi})={\tau(\mathcal{M}(\bm{\phi}_e(x),\bm{\phi}_o(\mathcal{S}_1)))},
\end{align}
where $\bm{\phi}_e$ as the first part of $\bm{\phi}$ represents the parameters of a sequence encoder ${\mathcal{R(\cdot)}}$, and $\bm{\phi}_o$ (the second part of $\bm{\phi}$)  denotes the embedding module for each candidate span in the span set $\mathcal{S}_1$. $\mathcal{M}(\cdot,\cdot)$ is a similarity function that measures the distance between two vectors. ${\tau}$ refers to the gumbel-softmax function~\cite{gumbel-softmax}, 
which guarantees sampling of the span with the largest probability, i.e., ${a_{out}^*}\sim p({a_{out}}|\bm{d};\bm{\phi})$, to be differentiable.
Implementation of the sampled action ${a_{out}^*}$ results in the selected span 
${s_{out}^*}$ 
to be substituted out.\\
\textbf{Parameterizing $p({a_{in}}|{a_{out}};\bm{d};\bm{\phi})$ for selection of spans to be substituted in}
The factors that govern the selection of a span to be swapped in from the whole span set $\mathcal{S}_2$ include the representations of (1) the span itself, (2) the input sentence ${x}$ for augmentation, and (3) the previously selected swap-out span ${s_{out}^*}$, so that those elusive spans that share the equivalence class with ${s_{out}^*}$ but contribute novel compositions via recombination with surroundings in $x$ are prioritized. 
Consequently, the probability distribution $p({a_{in}}|{a_{out}},\bm{d};\bm{\phi})$ over all $v$ candidate spans in $S_2$ follows,
\begin{align}
\label{eq:in_distribution}
 &\mathbf{c}=[\bm{\phi}_e(x);\bm{\phi}_{o}(s_{out}^*)]), \nonumber\\
 &p({a_{in}}|{a_{out}},\bm{d};\bm{\phi}) = \tau(\mathcal{M}(\bm{\phi}_{f}(\mathbf{c}),\bm{\phi}_i(\mathcal{S}_2))), 
\end{align}
where $\bm{\phi}_{f}$ and $\bm{\phi}_{i}$ altogether act as the third part of $\bm{\phi}$. Specifically, $\bm{\phi}_{i}$ is the embedding module for all spans in the span set $\mathcal{S}_2$ and $\bm{\phi}_{f}$ aligns the concatenated representation of the sentence and the swap-out span, i.e., $\bm{c}$, with $\bm{\phi}_i(\mathcal{S}_2)$ into the commensurable space.
Being consistent with the previous paragraph, we leverage the similarity function $\mathcal{M}(\cdot,\cdot)$ and the gumbel-softmax trick $\tau$ to sample ${a_{in}^*}\!\sim\! p({a_{in}}|{a_{out}^*},\bm{d};\bm{\phi})$.
It is noteworthy that we manually set the probability 
${a_{in}}\!\rightarrow\! {0}$ if ${\Pi(s_{in})\neq \Pi(s_{out}^*)}$ to excluse those potentially illegal synthesized examples.
The action ${a_{in}^*}$ finalizes 
the span ${s_{in}^*}$ 
to be substituted in.

\subsection{Training Procedures for L2S2}
Training L2S2 boils down to two alternating procedures: first, the generated examples by the L2S2 augmenter pass forward to train the down-stream neural sequence-to-sequence model parameterized by $\bm{\theta}$; second, the performance of the neural sequence model serves as feedback to update the upstream augmenter parameterized by $\bm{\phi}=\{\bm{\phi}_{e},\bm{\phi}_{o},\bm{\phi}_{i},\bm{\phi}_{f}\}$. \\
\textbf{Training objective for the seq-to-seq model} 
The 
objective of training the 
seq-to-seq model is to minimize the expected 
negative log-likelihood of producing the output sequence ${y_{gen}}$ from the input one ${x_{gen}}$ conditioned on the its parameters $\bm{\theta}$, i.e.,

\small
\begin{align}
\label{obj:theta}
\min_{\bm{\theta}}\mathcal{L}^{s}(\bm{\theta}) &=   {\min_{\bm{\theta}}{\mathbb{E}_{\bm{d}_{gen}\sim \mathcal{D}_{gen}}[-\log p(y_{gen}|x_{gen};\bm{\theta})]}}\nonumber \\
   & \approx \min_{\bm{\theta}}-\frac{1}{NT}\sum_{n=1}^{N}\sum_{t=1}^T\log p(y_{gen}^{n,t}|x_{gen}^{n,t};\bm{\theta}).
\end{align}
\normalsize
We would highlight that the empirical estimation samples over not only $N$ examples but also $T$ sequences of actions for each example, thus avoiding the randomness and high variance induced by the gumbel softmax trick. Thus, $(x_{gen}^{n,t},y_{gen}^{n,t})$ denotes a generated example from the $n$-th original training example by following the $t$-th sampled action sequence $(a^{n,t}_{out},a^{n,t}_{in})$. $\mathcal{D}_{gen}$ represents the distribution of all generated samples by the augmenter.\\
\textbf{Training objective for the L2S2 augmenter}
Our main purpose is to encourage the upstream L2S2 augmenter
to outweigh those challenging compositions by the elusive spans and novel surroundings. 
To achieve this goal, we evaluate the difficulty of a newly composed example $\bm{d}_{gen}$ by 
the feedback from the down-stream seq-to-seq model, i.e.,
the negative log-likelihood of predicting it; the larger the negative log-likelihood is, the more challenging the generated example is. 
Intuitively, we solve the following optimization problem to train the L2S2 augmenter to maximize the difficulty of synthesized examples.

\small
\begin{align}
\label{obj:phi}
& \max_{\bm{\phi}}\mathcal{L}^{a}(\bm{\phi}) =   {\max_{\bm{\phi}}{\mathbb{E}_{\bm{d}_{gen}\sim \mathcal{D}_{gen}}[-\log p(y_{gen}|x_{gen};\bm{\theta})]}}\nonumber \\
   & \approx \max_{\bm{\phi}}-\frac{1}{NT}\sum_{n=1}^{N}\sum_{t=1}^T p(\bm{d}_{gen}^{n,t}|\bm{d}^{n,t};\bm{\phi})\log p(y_{gen}^{n,t}|x_{gen}^{n,t};\bm{\theta}),
\end{align}
\normalsize
where $p(\bm{d}_{gen}^{n,t}|\bm{d}^{n,t};\bm{\phi})$ refers to the gumbel softmax probability distribution of the $t$-th sampled action sequence $(a^{n,t}_{out},a^{n,t}_{in})$ that translates $\bm{d}^{n,t}$ into $\bm{d}_{gen}^{n,t}$. 
To keep the L2S2 augmenter timely posted of the training state of the neural seq-to-seq model, we alternatingly optimize these two parts. We present the pseudo codes for training L2S2 in Alg.~\ref{algo:l2s2} in the Appendix.~\ref{append:l2s2_algorithm}.

\section{Experiments}

\begin{table*}[t]
\centering
\resizebox{\textwidth}{!}{
\begin{tabular}{lcc||ccc}
\hline
Method &Jump & Around Right& MCD1 & MCD2 & MCD3\\
\hline 
CGPS\footnotesize{~\cite{primsub}} &$98.8\%$\footnotesize{$\pm$ $1.4\%$} & $83.2\%$\footnotesize{$\pm$ $13.2\%$} &$1.2\%$\footnotesize{$\pm$ $1.0\%$} &$1.7\%$\footnotesize{$\pm$ $2.0\%$} &$0.6\%$\footnotesize{$\pm$ $0.3\%$}   \\
GECA+MAML\footnotesize{~\cite{meta-comp}}  & -- & -- & $58.9\%$\footnotesize{$\pm$ $6.4\%$}& $34.5\%$\footnotesize{$\pm$ $2.5\%$}& $12.3\%$\footnotesize{$\pm$ $4.9\%$} \\
Comp-IBT\footnotesize{~\cite{compibt}}  & $99.6\%$ & $37.8\%$ & $64.3\%$& $80.8\%$& $52.2\% $\\
T5-11B\footnotesize{~\cite{t5}}  & $98.3\%$ & $49.2\% $&$7.9\%$& $2.4\%$&$ 16.2\% $\\
\hline
\hline
LSTM  &$1.3\%$\footnotesize{$\pm$ $0.4\%$}& $10.2\%$\footnotesize{$\pm$ $4.6\%$} &$8.9\%$\footnotesize{$\pm$ $1.6\%$} &$11.9\%$\footnotesize{$\pm$ $9.4\%$}  & $6.0\%$\footnotesize{$\pm$ $0.9\%$}  \\
\hline 
+GECA\footnotesize{~\cite{geca}} &$95.2\%$\footnotesize{$\pm$ $8.0\%$} & $84.3\%$\footnotesize{$\pm$ $6.3\%$}&$23.4\%$\footnotesize{$\pm$ $9.1\%$} &$25.5\%$\footnotesize{$\pm$ $8.8\%$}  & $10.9\%$\footnotesize{$\pm$ $4.6\%$}  \\
+LexLearn\footnotesize{~\cite{lexlearn}} &$91.2\%$\footnotesize{$\pm$ $11.9\%$} & $95.3\%$\footnotesize{$\pm$$ 1.6\%$} & $12.5\%$\footnotesize{$\pm$ $2.0\%$} &$ 19.3\%$\footnotesize{$\pm$ $1.9\%$} & $ 11.6\%$\footnotesize{$\pm$ $0.9\%$}  \\
+LexSym\footnotesize{~\cite{lexsym}} &$100.0\%$\footnotesize{$\pm$ $0.0\%$} &$84.0\%$\footnotesize{$\pm$$ 7.1\%$} &$47.4\%$\footnotesize{$\pm$ $7.1\%$} &$30.8\%$\footnotesize{$\pm$ $8.4\%$}  & $13.7\%$\footnotesize{$\pm$ $3.6\%$}  \\
+Prim2PrimX+MET\footnotesize{~\cite{mutual}}  &$7.3\%$\footnotesize{$\pm$ $5.6\%$} & $97.6\%$\footnotesize{$\pm$ $1.0\%$}&$ 31.5\%$\footnotesize{$\pm$ $4.1\%$}&$ 33.5\%$\footnotesize{$\pm$ $2.7\%$}&  $11.6\%$\footnotesize{$\pm$ $1.0\%$} \\
+GECA+MAML\footnotesize{~\cite{meta-comp}}  & $95.8\%$\footnotesize{$\pm$ $6.9\%$}& $86.2\%$\footnotesize{$\pm$ $5.6\%$}& $28.2\%$\footnotesize{$\pm$ $9.6\%$}& $31.8\%$\footnotesize{$\pm$ $8.5\%$}& $11.2\%$\footnotesize{$\pm$ $4.2\%$} \\
\hline 
+SpanSub \footnotesize{(\textbf{Ours})} &{$\bm{100.0\%}$}\footnotesize{$\pm$ $0.0\%$} &$99.9\%$\footnotesize{$\pm$$ 0.1\%$} &$63.4\%$\footnotesize{$\pm$ $13.1\%$} &$72.9\%$\footnotesize{$\pm$ $10.1\%$}  & $74.0\%$\footnotesize{$\pm$ $10.2\%$}  \\
+SpanSub+L2S2 \footnotesize{(\textbf{Ours})}  &$\bm{100.0\%}$\footnotesize{$\pm$ $0.0\%$} &$\bm{100.0\%}$\footnotesize{$\pm$ $0.0\%$} &$\bm{67.4\%}$\footnotesize{$\pm$ $12.1\%$} &$\bm{73.0\%}$\footnotesize{$\pm$ $10.1\%$}  & $\bm{80.2\%}$\footnotesize{$\pm$ $1.8\%$}  \\
\hline 
\end{tabular}
}
\caption{
Test accuracy on SCAN Jump, Around Right and MCD splits.
}
\label{tab:scan_exps}
\end{table*}

\begin{table}[t]
\centering
\resizebox{0.4\textwidth}{!}{
\begin{tabular}{lc}
\hline
Method & COGS\\
\hline
MAML\scriptsize{~\cite{meta-comp}}  &$  64.1\% \tiny{\pm 3.2\%}  $\\
IR-Transformer\scriptsize{~\cite{IR-transf}}   & $  78.4\% $\\
Roberta+Dangle\scriptsize{~\cite{dangle}}   & $   87.6\%$ \\
T5-Base\scriptsize{~\cite{t5}}   &  $  85.9\% $\\
\hline 
\hline
LSTM  & $ 55.4\% \tiny{\pm 4.2\%} $\\
\hline 
+GECA\scriptsize{~\cite{geca}}   &  $ 48.0\% $\tiny{$\pm 5.0\%$} \\
+LexLearn\scriptsize{~\cite{lexlearn}}  & $ 82.0\%$ \tiny{$\pm 0.0\%$} \\
+LexSym\scriptsize{~\cite{lexsym}}  &  $ 81.4\% $\tiny{$\pm 0.5\%$}  \\
+Prim2PrimX+MET\scriptsize{~\cite{mutual}}  &$ 81.1\% $\tiny{$\pm 1.0\%$} \\
\hline 
+SpanSub \footnotesize{(\textbf{Ours})}  & $91.8\% $\tiny{$\pm 0.1\%$} \\
+SpanSub+L2S2 \footnotesize{(\textbf{Ours})} & $\bm{92.3\%}$\tiny{$\pm 0.2\%$}  \\
\hline 
\end{tabular}
}
\caption{
Overall test accuracy on COGS dataset.
}
\label{tab:cogs_exps}
\end{table}

\begin{table}[h!]
\centering
\resizebox{0.4\textwidth}{!}{
\begin{tabular}{lc|c}
\hline
Method & question & query \\
\hline
SpanParse\scriptsize{~\cite{spanparse}}& $78.9\%$ & $76.3\%$\\
\hline 
\textbf{LSTM} &$ 75.2\%$ & $58.6\%$\\
+GECA\scriptsize{~\cite{geca}} & $76.8\%$ &  $60.6\%   $\\
+LexSym\scriptsize{~\cite{lexsym}} &$81.6\%$ & $80.2\% $   \\
+SUBS\scriptsize{~\cite{subs}}  &$80.5\%$ &  $77.7\%  $ \\
+SpanSub \footnotesize{(\textbf{Ours})}&$ \bm{82.4\%}$ & $\bm{81.4\%} $\\
\hline 
\hline
\textbf{BART}\scriptsize{~\cite{bart}} &$90.2\%$ & $71.9\%$\\
+GECA\scriptsize{~\cite{geca}} & $87.9\%$ &   $83.0\%  $\\
+LexSym\scriptsize{~\cite{lexsym}} &$90.2\%$ & $87.7\%  $  \\
+SUBS\scriptsize{~\cite{subs}} & $\bm{91.8\%}$ &  $88.3\%   $\\
+SpanSub \footnotesize{(\textbf{Ours})} & $90.6\%$ & $\bm{89.5\%} $\\
\hline 
\end{tabular}
}
\caption{
Test accuracy on GeoQuery question (i.i.d.) and query (compositional) splits.
}
\label{tab:geoquery_exps}
\end{table}

\subsection{Datasets and Splits}
We evaluate our proposed methods on the following three popular and representative semantic parsing benchmarks which target for challenging the compositional generalization capacity of neural sequence models. These benchmarks contain not only synthetic evaluations deliberately designed for diverse categories of systematic generalization but also non-synthetic ones additionally requiring capabilities of neural models in handling natural language variations~\cite{nqg}. More detailed descriptions of these datasets can be found in Appendix~\ref{append:data}.\\
\textbf{SCAN} Introduced by~\cite{scan}, SCAN contains a large set of synthetic paired sequences whose input is a sequence of navigation commands in natural language and output is the corresponding action sequence. Following previous works~\cite{geca,lexlearn,mutual}, we evaluate our methods on the two splits of \emph{\textbf{jump}} (designed for evaluating a novel combination of a seen primitive, i.e., \emph{jump}, and other seen surroundings) and \emph{\textbf{around right}} (designed for evaluating a novel compositional rule).
Notably, we also consider the more complex and challenging Maximum Compound Divergence (MCD) splits of SCAN established in~\cite{mcd}, which distinguish the compound 
distributions of the training and the testing set as sharply as possible.\\
\textbf{COGS} Another synthetic COGS dataset~\cite{cogs} contains 24,155 pairs of English sentences and their corresponding logical forms. COGS contains a variety of systematic linguistic abstractions (e.g., active $\rightarrow$ passive, nominative $\rightarrow$ accusative and transtive verbs $\rightarrow$ intranstive verbs), thus reflecting compositionality of natural utterance. 
It is noteworthy that COGS with its testing data categorized into 21 classes by the compositional generalization type supports fine-grained evaluations.\\
\textbf{GeoQuery} 
The non-synthetic dataset of GeoQeury~\cite{geoquery} collects 880 anthropogenic questions regarding the US geography (e.g., "what states does the mississippi run through ?") paired with their corresponding database query statements (e.g., "answer ( state ( traverse\_1 ( riverid ( mississippi ) ) ) )"). Following~\cite{spanparse,subs}, we also adopt the FunQl formalism of GeoQuery introduced by~\cite{funql} and evaluate our methods on 
the compositional template split (\emph{\textbf{query}} split) from~\cite{template-split} where the output query statement templates 
of the training and testing set are disjoint and the \emph{i.i.d.} split (\emph{\textbf{question}} split) where training set and testing set are randomly separated from the whole dataset.

\subsection{Experimental Setup}
\textbf{Baselines}
We compare our methods with the following prior state-of-the-art baselines for compositional generalization.
(1) Data augmentation methods: GECA~\cite{geca} and LexSym~\cite{lexsym} on all the three benchmarks, Prim2PrimX+MET~\cite{mutual} which is a data augmentation methods further boosted by mutual exclusive training on SCAN and COGS, and SUBS~\cite{subs} as the current state-of-the-art on GeoQuery. Besides, we additionally compare our methods with GECA+MAML~\cite{meta-comp}(boost GECA with meta-learning) and Comp-IBT~\cite{compibt} which is also a data augmentation method requiring to access 30\% testing inputs and outputs in advance.
(2) Methods that incorporate the alignment of tokens or substructures: LexLearn~\cite{lexlearn} on SCAN and COGS,
IR-Transformer~\cite{IR-transf} on COGS, as well as SpanParse~\cite{spanparse} on GeoQuery.
(3) Methods that design specialized architectures: CGPS~\cite{primsub} on SCAN and Roberta+Dangle~\cite{dangle} on COGS.
(4) We also report the results on SCAN and COGS from powerful pretrained T5~\cite{t5} as reference.\\
\textbf{Base Models}
In alignment with the previous works~\cite{geca,lexlearn,lexsym}, we adopt the 
LSTM-based seq-to-seq model~\cite{seq2seq} with the attention~\cite{attention} and copy~\cite{copy} mechanisms as our base model on the SCAN and COGS benchmarks. 
For the non-synthetic dataset of GeoQuery, we follow
SpanParse~\cite{spanparse} and SUBS~\cite{subs} by using not only LSTM but also a more capable pre-trained language model BART~\cite{bart} as our base models. 
Detailed experimental settings 
are available in Appendix~\ref{append:model}.\\
\textbf{Evaluation Metric}
Grounded on the semantic parsing task, we adopt the evaluation metric of exact-match accuracy in all of our experiments. 
\subsection{Main Results}
\label{sec:main_result}
The results of our experiments on SCAN, COGS and GeoQuery benchmarks are shown in Table \ref{tab:scan_exps}, Table ~\ref{tab:cogs_exps} and Table ~\ref{tab:geoquery_exps} respectively.
Note that \textbf{"+SpanSub"} means that we directly use SpanSub to generate additional training data and train our base models on the original training data and the additional training data generated by SpanSub as well; \textbf{"+SpanSub+L2S2"} means that we (1): firstly augment the original training data with additionally generated data using SpanSub, (2): train the L2S2 framework (using Algorithm~\ref{algo:l2s2}) on the augmented training data, and (3): get the trained base models from the L2S2 framework. We run each experiment on the 5 different seeds and report the mean and the standard deviation. We also do ablation studies and control experiments (in Appendix.~\ref{appendix:ablation}) to separately verify the effectiveness of SpanSub and L2S2 and their combination. \\
\textbf{SCAN Results}
On all of the 5 splits (jump, around right, MCD1, MCD2 and MCD3) which we study in the SCAN benchmarks, SpanSub and the combination of it and L2S2 both lead to significant improvements for our base models.
For easier/classic \emph{\textbf{jump}} and \emph{\textbf{around right}} splits, the performance of our base model improves to solving these two tasks completely. 
For more chanllenging \emph{\textbf{MCD}} splits, when we leverage SpanSub to generate additional training data for our base model, the performance of it improves around 64\% on average. Moreover, the adoption of L2S2 further boosts the performance by at most 6.2\% on the basis of only using SpanSub.
Using our methods obviously outperforms using the majority of other baseline methods, except for Comp-IBT on MCD2 split. Nonetheless, Comp-IBT requires to access 30\% inputs and outputs in the testing set, so it is not directly comparable with ours.\\
\textbf{COGS Results}
On COGS task, the performance of our base model(LSTM) increase from 55.4\% to 91.8\% when we use SpanSub to generate additional training data for it. SpanSub has approximately 10\% lead compared with our baseline methods (LexLearn, LexSym, Prim2PrimX+MET) implemented on the same base model. Even compared with methods that leverage powerful pretrained models (e.g., Roberta+Dangle and T5-Base), LSTM+SpanSub still has some advantages. Furthermore, through adopting L2S2 on the basis of SpanSub, we can improve the performance of our base model from 91.8\% to 92.3\%.\\
\textbf{GeoQuery Results}
On the compositional template \emph{\textbf{query}} split, SpanSub leads to substantial and consistent improvement over other baseline data augmentation methods (GECA, LexSym and SUBS) on both of implementations based on LSTM and BART, achieving new state-of-the-art results (pushing forward the previously state-of-the-art results by 1.2\%). 
As for the \emph{i.i.d} \emph{\textbf{question}} split, SpanSub still has advantages over baseline methods when based on LSTM model. When we adopt BART as our base model, SpanSub boosts the performance of BART by 0.4\% which is ahead of GECA and LexSym, falling behind SUBS.

\subsection{Analysis and Discussion}
\begin{figure}
\centering 
\includegraphics[width=0.47\textwidth]{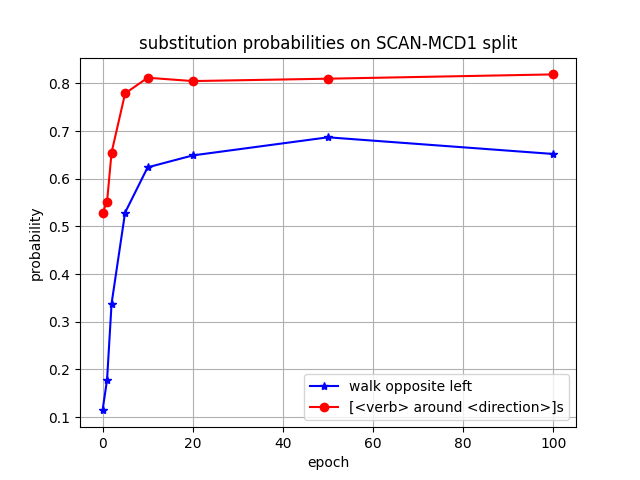}
\caption{The variation curves of substitution probabilities with the training process going on. Given an training example "run right thrice after walk opposite left twice",  The blue curve represents the variation curve of probabilities of swapping "walk opposite left" out and the red curve represents the variation curve of probabilities of swapping spans like "<verb> around <direction>s" in.}
\label{figure:sub_prob}
\end{figure}

In this section, we aim to further answer the following four questions:
\begin{itemize}
    \item Does the SpanSub help with fully exploring of augmentation space as supposed in Section~\ref{sec:intro}?
    \item Does the L2S2 learn to realize the hardness-aware automatic data augmentation as supposed in Section~\ref{sec:intro}?
    \item Ablation Studies and Control Experiments: Do the L2S2 and the SpanSub separately help with compositional generalization? Can their combination further improve generalization capactiy? Does the up-stream learnable augmentation module play an necessary role?
    \item Can the proposed L2S2 methods generalize to more types of down-stream neural sequence models (other than LSTM-based models, e.g., Transformers~\cite{transformer})?
\end{itemize}
\textbf{Analysis of performances with SpanSub}
\begin{table}
\centering
\resizebox{0.4\textwidth}{!}{
\begin{tabular}{l|c|c|c|c}
\hline
Method & lex & s1 & s2 & s3 \\
\hline
\footnotesize{LSTM} & \footnotesize{$69.3\%$} & \footnotesize{$0.0\%$} & \footnotesize{$0.0\%$} & \footnotesize{$0.9\%$}\\
\footnotesize{+LexSym} & \footnotesize{$95.3\%$} & \footnotesize{$0.0\%$} & \footnotesize{$0.0\%$} & \footnotesize{$0.7\%$}\\ 
\hline
\footnotesize{+SpanSub} & \footnotesize{$99.1\%$} & \footnotesize{$91.8\%$} & \footnotesize{$45.0\%$} & \footnotesize{$7.2\%$}\\
\footnotesize{+SpanSub+L2S2} & \footnotesize{$99.4\%$} & \footnotesize{$93.7\%$} & \footnotesize{$45.1\%$}& \footnotesize{$10.7\%$}\\
\hline
\end{tabular}
}
\caption{
Test accuracy of different generalization types in COGS task. "lex" refers to lexical generalization test; "s1","s2" and "s3" refer to "obj\_pp\_to\_subj\_pp", "pp\_recursion", "cp\_recursion" respectively, which are 3 different types of structural generalization tests.
}
\label{tab:cogs_type}
\end{table}
To further analyze the improvement of performance brought by SpanSub and L2S2, we break down the the performance on COGS task into four different part, including lexical generalization performance and three different types of structural generalization performances. Results are shown in Table~\ref{tab:cogs_type}.
Compared with LexSym, which only enable single-grained substitutions (i.e., substituting for single words), we find that SpanSub can not only improve generalization on testing cases of different structural types, but also further boost the lexical level generalization.\\
\textbf{Analysis of performances with L2S2}
For results on SCAN(MCDs) tasks:
We investigate the concrete substitution probabilities generated by L2S2 augmentor on MCD1 (where the complex concept "<verb> around <direction>" never co-occur with "twice" in the training set) split of SCAN task (training only with L2S2 framework). Given an example "run right thrice after walk opposite left twice", we keep on observing the probabilities of L2S2 augmentor selecting the span "walk opposite left" to be swapped out and selecting the spans like "<verb> around <direction>" to be swapped in, with the training process going on. The results are shown in Fig~\ref{figure:sub_prob}.\footnote{In this figure we count "epoch" (x-axis) after the end of the warm-up stage.}
As the training process goes on, L2S2 augmentor learns to compose spans like "<verb> around <direction>" and novel surrounding "twice". This exactly verify our hypothesis that L2S2 framework can automatically learn to put high value on the compositions of elusive concepts and novel surroundings. As a comparison with imbalanced prediction error rates shown in Fig~\ref{figure:intro}(d), we put the results of additionally using L2S2 and RandS2 (which is the controlled version of L2S2, by substituting the learned parameters in the L2S2 with random ones.) in Table~\ref{tab:w_a_r_err_dec}. We can conclude that L2S2 can effectively help with the performance of down-stream neural seq-to-seq models on the prediction of harder examples.\footnote{Note that Fig~\ref{figure:intro}(d) shows the results on SCAN-MCD1, and Table~\ref{tab:w_a_r_err_dec} shows the results on SCAN-MCD3. This slight mismatch does not change our conclusion here.}

\begin{figure}
\centering 
\includegraphics[width=0.47\textwidth]{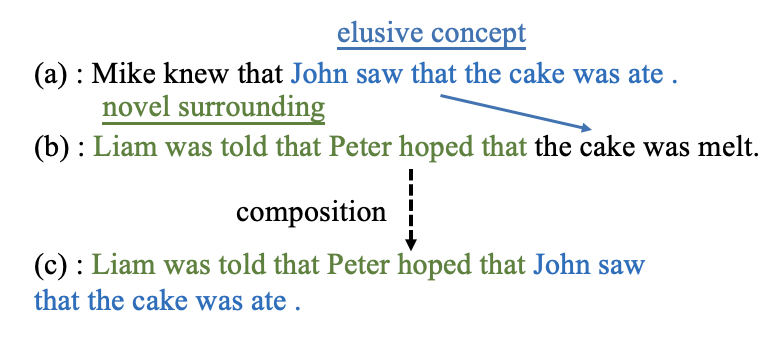}
\caption{A composition that helps to improve "cp\_recursion" generalization in SpanSub. The composition of "John saw that the cake was ate" and "Liam was told that Peter was hoped that" results in examples with deeper recursion of \textbf{that}-structure.}
\label{figure:cogs_case}
\end{figure}

For results on the COGS task: 
as shown in Table~\ref{tab:cogs_type}, we find that the utilization of L2S2 framework training can help SpanSub better generalize on testing cases of "cp\_recursion" type. As shown in Fig~\ref{figure:cogs_case}, in SpanSub, "cp\_recursion" type generalization cases require the compositions of concepts of sentential complements (e.g., "John knew \textbf{that} the cake was ate .") and novel surroundings (with deep recursion of \textbf{that}-structure). L2S2 framework training improves SpanSub on "cp\_recursion" generalization through encouraging such compositions. \\
\textbf{Ablation Study}
Except for the performance analysis provided above, we also do ablation study and control experiments to separately verify the effectiveness of SpanSub, L2S2 and their combination. Due to the page limit, our detailed experiment setting and results are shown in Table~\ref{tab:ablation_exps} in Appendix~\ref{append:experiments}. \\
\textbf{Generalizing L2S2 to more based models}
Since we claim that our proposed L2S2 method is model-agnostic, here we generalize it to three different kind of base models\footnote{here the term "base model" refers to down-stream neural seq-to-seq models in Fig~\ref{algo:l2s2}.}: one-layer LSTM used in ~\cite{geca}, two-layer LSTM used in ~\cite{lexlearn} and Transformer used in ~\cite{mutual}. The experiments results are shown in Table~\ref{tab:different_archs_exps} in Appendix~\ref{append:experiments}.

\section{Conclusion}
In this paper, (1) we present a novel substitution-based compositional data augmentation scheme, SpanSub, to enable multi-grained compositions of substantial substructures in the whole training set and (2) we introduce an online, optimizable and model-agnostic L2S2 framework containing a L2S2 augmentor which automatically learn the span substitution probabilities to put high values on those challenging compositions of elusive spans and novel surroundings, and thus further boost the systematic generalization ability of down-stream nerual sequence models especially on those hard-to-learn compositions. Empirical results demonstrate the effectiveness and superiority of SpanSub, L2SS and their combination.
\section{Limitations}
The techniques in SpanSub are constructed on the basis prior works of extracting span alignments and clustering words in the training data according to their syntactic role. There is no generic solution for these problem applicable for all of the datasets (this is mainly because the output formats and structures are diverse) at present, which requires users to spend efforts looking for preprocessing techniques applicable for their own datasets. However, the methodology of the proposed SpanSub is rather general to many different datasets and tasks (e.g., Semantic Parsing and Machine Translation). Besides, although we define eligible spans to try to alleviate additionally introducing noisy augmented data, our experiment result on GeoQuery (i.i.d. split) shows that SpanSub can still slightly hurt generalization performance (in comparison with other state-of-the-art methods). Hence we regard that relieving the potentially negative influence of noisy augmentation is important to further improve this work.
\section{Acknowledgement}
We sincerely thank the anonymous reviewers for giving useful feedback and constructive suggestions to the initial version of the paper. This work  was supported by grants from the National Key R\&D Program of China (No. 2021ZD0111801) and the National Natural Science Foundation of China (No. 62022077).

\bibliography{acl2023}
\bibliographystyle{acl_natbib}

\appendix

\section{Datasets and Preprocessing}
\label{append:data}
\subsection{Datasets}
\label{append:data_intro}
\paragraph{SCAN} Introduced by~\cite{scan}, SCAN contains a large set of synthetic paired sequences whose input is a sequence of navigation commands in natural language and output is the corresponding action sequence. Following previous works~\cite{geca,lexlearn,mutual}, we evaluate our methods on the two splits of \emph{\textbf{jump}} (designed for evaluating a novel combination of a seen primitive, i.e., \emph{jump}, and other seen surroundings) and \emph{\textbf{around right}} (designed for evaluating a novel compositional rule).
Notably, we also consider the more complex and challenging Maximum Compound Divergence (MCD) splits of SCAN established in~\cite{mcd}, which distinguish the compound 
distributions of the training and the testing set as sharply as possible.
\paragraph{COGS} Another synthetic COGS dataset~\cite{cogs} contains 24,155 pairs of English sentences and their corresponding logical forms. COGS contains a variety of systematic linguistic abstractions (e.g., active $\rightarrow$ passive, nominative $\rightarrow$ accusative and transtive verbs $\rightarrow$ intranstive verbs), thus reflecting compositionality of natural utterance. 
It is noteworthy that COGS with its testing data categorized into 21 classes by the compositional generalization type supports fine-grained evaluations.
\paragraph{GeoQuery} 
The non-synthetic dataset of GeoQeury~\cite{geoquery} collects 880 anthropogenic questions regarding the US geography (e.g., "what states does the mississippi run through ?") paired with their corresponding database query statements (e.g., "answer ( state ( traverse\_1 ( riverid ( mississippi ) ) ) )"). Following~\cite{spanparse,subs}, we also adopt the FunQl formalism of GeoQuery introduced by~\cite{funql} and evaluate our methods on 
the compositional template split (\emph{\textbf{query}} split) from~\cite{template-split} where the output query statement templates 
of the training and testing set are disjoint and the \emph{i.i.d.} split (\emph{\textbf{question}} split) where training set and testing set are randomly separated from the whole dataset.

We provide examples of the above three datasets as follows for readers' reference:
\begin{lstlisting}
// a SCAN example 
scan["input"] = 
    "walk around right twice and jump left thrice"
scan["target"] = 
    "TR W TR W TR W TR W TR W TR W
    TR W TR W TL J TL J TL J"
// a COGS example
cogs["input"] = 
    "Amelia gave Emma a strawberry ."
cogs["target"] =
    "give . agent ( x _ 1 , Amelia ) AND give . recipient ( x _ 1 , Emma ) 
    AND give . theme ( x _ 1 , x _ 4 ) AND strawberry ( x _ 4 )"
// a GeoQuery example
geoquery["input"] = 
    "what is the tallest mountain in america ?"
geoquery["target"] = 
    "answer ( highest ( mountain ( loc_2 ( countryid ( 'usa' ) ) ) ) )"
\end{lstlisting}
\subsection{Proprocessing of Datasets}
\label{append:data_preprocess}
\paragraph{Extraction of span alignments}
For SCAN dataset, since there is no off-the-shelf technique to map sequential data in SCAN dataset to tree-form, we slightly the modify algorithm SimpleAlign from ~\cite{lexlearn} to extract consecutive span alignments for our experiments on SCAN. We denote the input sequence as $x$, the output sequence as $y$, the span, which is going to be extracted from the input sequence, as $v$ and its counterpart in the output sequence as $w$. Basically, we extract a pair of span alignment $(v,w)$ following the maximally restrictive criterion:
\begin{equation}
\begin{split}
& nec.(v,w)=\forall xy. (w\in y)\rightarrow (v\in x)\\
& suff.(v,w)=\forall xy. (v\in x)\rightarrow (w\in y)\\
& C_1(v,w) = nec.(v,w) \land  suff.(v,w)\\
\end{split}
\label{equ:scan_span}
\end{equation}
Both $v$ and $w$ are supposed to be consecutive fragments in the input sequence and output sequence respectively. \\
We additionally apply appropriate relaxations on the top of criterion(~\ref{equ:scan_span}) to enable the extraction of more spans: we tolerate many-to-one mapping and one-to-many mapping to some extent to avoid discarding of "<verb>s around <direction>s" and "<verb>s <direction>s"(e.g., both of interpretations of "walk around right" and "walk right" cover "TR W"). 
Besides, we manually set the maximum number of words in $v$ to 3 and the maximum number of words in $w$ to 8. 

\begin{figure}[t]
\centering 
\includegraphics[width=0.47\textwidth]{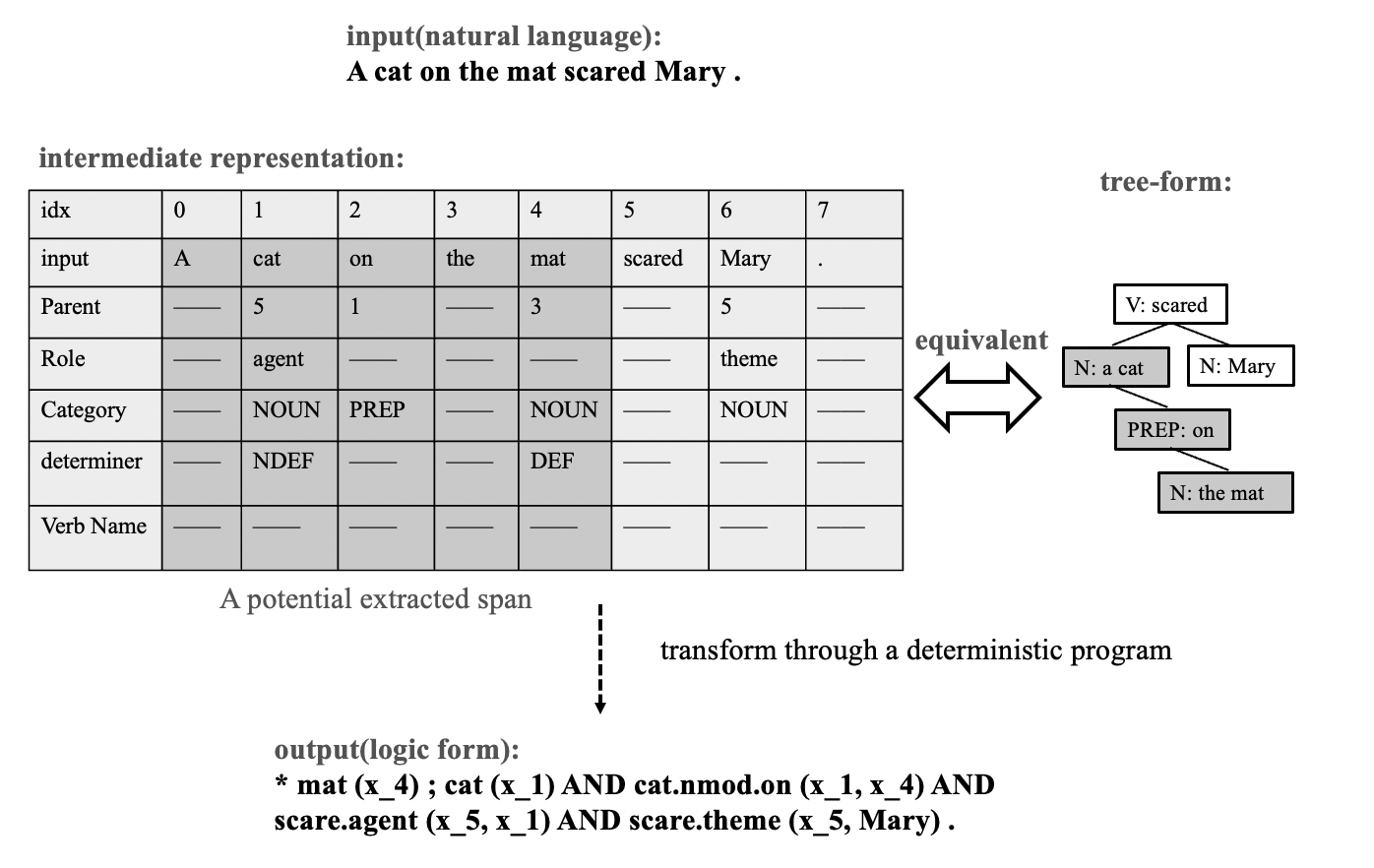}
\caption{An instance for an intermediate representation, its corresponding tree-form and a potential extracted span for COGS.}
\label{figure:cogs_align_instance}
\end{figure}

For COGS, we directly use the intermediate representation from ~\cite{IR-transf}. An instance of intermediate representation is shown in Fig~\ref{figure:cogs_align_instance}.
We search for every consecutive fragments in the intermediate presentations of COGS to extract eligible spans according to Definition~\ref{define:tree}. The naive implementation of the above search algorithm has the time complexity of $\mathcal{O}(n\cdot m^3)$, where $n$ is the number of sentences in the training set and $m$ is the maximal length of a single sentence in the training set.

For GeoQuery, following ~\cite{subs}, we directly adopt the span trees (\emph{gold trees}) extracted and aligned by ~\cite{spanparse}. And we refer the readers to get more detailed information about how to construct such span trees from the original paper~\cite{spanparse}.\\
Note that we slightly correct several denotations in the original \emph{gold trees} from ~\cite{spanparse}, for they are slightly differing from the ground-truth. To clarify it, we put an example of modification here (given that the others are similar, we do not present the others here):
\begin{lstlisting}
geoquery["input"] = 
    "what is the population of washington dc ?"
geoquery["program"] = 
    "answer ( population_1 ( cityid ( 'washington', 'dc' ) ) )"
// the original gold_spans
geoquery["gold_spans"] = 
    {"span": [5, 5], "type": "cityid#'washington'"} 
// after correction
geoquery["gold_spans"] = 
    {"span": [5, 6], "type": "cityid#'washington'"} 
    // this is just one of the spans
    // washington dc is the capital city of USA; 
    // washington is a state of USA;
\end{lstlisting}
To ensure a fair comparison with previous substitution-based data augmentation methods~\cite{lexsym,subs}, we rerun their methods on the modified \emph{gold trees}.
\begin{figure}[t]
\centering 
\includegraphics[width=0.47\textwidth]{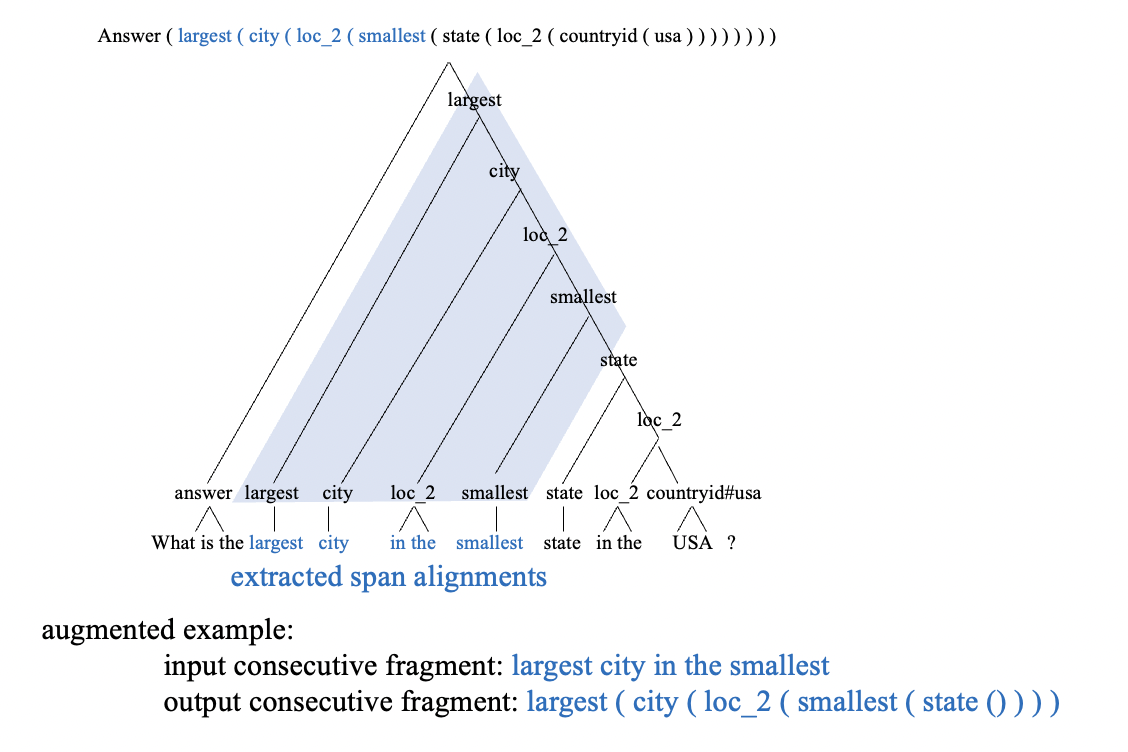}
\caption{An instance for a constructed span tree and extracting a consecutive span from the span tree.}
\label{figure:geoquery_instance}
\end{figure}
\paragraph{Inferring the equivalence class of words}
For COGS, we directly leverage the information in the intermediate representations to infer the equivalence class of the words (e.g., NOUN, VERB or PREP). For SCAN and GeoQuery, we use the technique of inferring the types of words form ~\cite{lexsym}, which cluster the words according to their shared contexts in the training set.\\
For GeoQuery, we additionally adopt context2vec methods~\cite{melamud-etal-2016-context2vec} (where we train a simple one-layer LSTM-based mask-reconstruction model) to boost the exploration of potentially syntactically-equivalent words (i.e., candidates to fill in the masked blank). We put the final result of word-clustering on GeoQuery here as follows:(We cluster the words in the target side)\\
\begin{lstlisting}
/*
word clustering result for GeoQuery: 
words not included are not syntactically 
equivalent to any other words
*/
cluster1 = ['highest','major','largest',
            'smallest','shortest','lowest',
            'longest']
cluster2 = ['mountain','state','city',
            'river','place','lake']
cluster3 = ['loc_2','traverse_2']
cluster4 = ['countryid','cityid','stateid',
            'placeid']
cluster5 = ['traverse_1','loc_1','capital_2']
cluster6 = ['largest_one','smallest_one']
cluster7 = ['area_1','density_1','population_1']
cluster8 = ['size','high_point_1']
cluster9 = ['most','fewest']
\end{lstlisting}
\section{Training Details and Hyper-parameter Selection of Algorithms}
\label{append:model}
In this section, we detailedly describe the training details of our models in our framework(up-stream L2S2 Augmentor and down-stream neural seq-to-seq model) and the selection of hyper-parameters in our Algorithms(SpanSub and L2S2). 
\subsection{L2S2 Augmentor}
For both of SCAN and COGS experiments, we use an two layer bidirectional LSTM (with 128 hidden units and an embedding size of 128, a dropout rate of 0.5) as our sequence encoder. We separately use an embedding layer with an embedding size of 512 for the embedding module for spans to be swapped out and another embedding layer with an embedding size of 512 for the embedding module for spans to be swapped in. We use (cosine-similarity$\cdot$2) $ \in [-2,2]$ as all of our similarity functions in L2S2 augmentor. We set all of the temperatures for gumbel-softmax sampling in L2S2 augmentor to 1.
Besides, we use a Adam optimizer~\cite{Kingma2014AdamAM} to optimize our L2S2 augmentor with an learning rate of 1e-3.
The above hyper-parameters are commonly used for LSTM-based models in NLP community and hence we did not spend extra efforts to tune them in our experiments.
\label{append:model_augmentor}
\subsection{Neural Seq-to-Seq Models}
\label{append:model_parser}
We keep this part of hyper-parameters aligned with previous baselines.
For \emph{jump} and \emph{around right} splits of SCAN and COGS experiments, we keep the hyperparameters of our LSTM in align with \cite{lexlearn, lexsym, mutual}. We use a 2-layer encoder-decoder LSTM (with attention~\cite{attention} and copy~\cite{copy} mechanisms) with 512 hidden units and an embedding size of 512, a droupout rate of 0.4. For \emph{MCD}1, \emph{MCD}2 and \emph{MCD}3 splits of SCAN experiments, the hyperparameters of our LSTM are adopted form \cite{geca}. We use a 1-layer bidirectional encoder-decoder LSTM (with attention and copy mechanisms) with 512 hidden units and an embedding size of 64, a droupout rate of 0.5. For all of these above experiments, we train our model with an Adam optimizer with an initial learning rate of 1e-3. We use an ReduceLROnPlateau scheduler (implemented in PyTorch) with a scale factor of 0.5 to automatically reduce our learning rate. We set all of the batch size to 128.

For GeoQuery tasks, in align with SUBS~\cite{subs}, we also directly use OpenNMT~\cite{opennmt} to implement our LSTM-based model with attention and copy mechanisms and we utilize fairseq~\cite{fairseq} to implement our BART-based model. For LSTM-based experiments, we use one-layer bidirectional LSTM in the encoder side and one-layer unidirectional LSTM in the decoder side. We use dropout with a rate of 0.5 and Adam optimizer with a learning rate of 1e-3. We use MLP attention and directly use the attention scores as copying scores and we set the batch size for experiments based on LSTM to 64. For BART-based experiments, we use BART-base models updated by Adam optimizer with a learning rate of 1e-5. We set the rate for both dropout and attention dropout to 0.1 and we use label smoothing with a rate of 0.1. We set the batch size for all of the experiments based on BART to 1024 tokens. Besides, we set the rate of the weight-decay to 0.01. 
\subsection{Hyper-parameters in SpanSub(Algorithm~\ref{algo:spansub})}
For \emph{jump} and \emph{around right} splits of SCAN and GeoQuery experiments, we set the iterative depth $K$ in SpanSub augmentation scheme to 1. For \emph{MCD} splits of SCAN experiments, we set the iterative depth $K$ in SpanSub augmentation scheme to 2. For COGS experiments, we set the iterative depth $K$ in SpanSub augmentation scheme to 4. For SCAN experiments, we set the number of generated examples $N$ (without de-duplicating) to 1e5. For COGS experiments, we set the number of generated examples $N$ (without de-duplicating) to 4e5. For GeoQuery experiments, we simply searching for every potential augmentations in the training set (because the training set for GeoQuery contains merely 519 examples, we try to make the best use of each example.), and the size of augmented set is shown in Table~\ref{tab:max_aug_num}. Following ~\cite{jia_liang, qiu_csl}, we also ensure approximately equal number of the original examples and the augmented examples being used for training in SpanSub experiments, giving consideration to both of i.i.d. generalization and compositional generalization.

We decide the iterative depth $K$ through observing that from which iteration there are nearly no more novel data generated. For $N$, we simply set a number which is large enough compared with the size of the original dataset, and then we de-duplicate the augmented dataset. 
\subsection{Hyper-parameters in Training L2S2 framework(Algorithm~\ref{algo:l2s2})}
One crucial hyper-parameter in Training L2S2 framework is the warm-up epochs / update steps. In most cases, we need to set an appropriate value to warm-up update steps to guarantee the down-stream sequence model to be fully aware of the distribution (hardness) of the original training examples while not over-fit to them.
For most of our experiments(\emph{jump}, \emph{around right}, \emph{MCD1} and \emph{MCD2} splits of SCAN experiments, COGS experiments), we set the warm-up epoch to 5, and then we alternatively train the up-stream module and down-stream module in the L2S2 framework to 150 epochs in total. For \emph{MCD2} split of SCAN experiments, we first train our neural seq-to-seq model for 80 epochs, and then we alternatively train the up-stream L2S2 augmentor and the down-stream neural seq-to-seq model for 70 epochs\footnote{In our initial experiments, we found that L2S2 method only slightly works on the \emph{MCD2} split of SCAN dataset when using 1 layer LSTM-based model as the down-stream sequence model. However, in the following experiments in Table~\ref{tab:different_archs_exps}, we found that it works well on other 2 down-stream sequence models (we set warm-up epoch number to 5 for other down-stream seq-to-seq models).}.
For experiments with L2S2 framework, we set the number of sampled actions $T$ for each example to 4. All of this part of hyper-parameters are decided by cross-validation.
\paragraph{Other Training Details}
We conduct all of our experiments on NVIDIA GeForce RTX2080Ti GPUs. For \emph{jump} and \emph{around right} splits of SCAN, COGS and GeoQuery, we select our model for testing with the best development accuracy.
For all \emph{MCD} splits of SCAN, we use the train/dev/test splits from the original paper~\cite{mcd}\footnote{The official github repo is \url{https://github.com/google-research/google-research/tree/master/cfq\#scan-mcd-splits}, and one can download the dataset from \url{https://storage.cloud.google.com/cfq_dataset/scan-splits.tar.gz}}, we also select our model for testing with the best accuracy on dev set.
\section{Definitions and Algorithms}
In this section, we mainly describe the pseudo-code of SpanSub and L2S2, and the formal description of the term "span".
\subsection{SpanSub}
Different from ~\cite{subs}, we extract any consecutive fragments as our spans. An instance for the constructed span tree and extracting a consecutive span from the span tree is shown in Fig~\ref{figure:geoquery_instance}. And we give the formal description of the term "span" used throughout this paper.
\label{append:define_spansub}
\begin{myDef}
\label{define:tree}
(\textbf{Eligible Span}) Given a sentence or a program sequence ${S}$ = ${[e_0,\ e_1,\ ...,\ e_n]}$, there exists one and only one multi-way tree ${T}$ corresponding to ${S}$, the in-order traversal sequence\footnote{In our case in-order traversal of a multi-way tree is to traverse the most left child, traverse the root node and then traverse left childs from right to left in order.} ${\Lambda}$ of which is ${v_0\rightarrow v_1\rightarrow ...\rightarrow v_n}$ (node ${v_i}$ corresponds to token ${e_i}$, $0 \leq i \leq n$). Any span ${S'}$ = ${[e_p, e_{p+1}, ..., e_{p+k}]}\subseteq$ ${S}$, where $0 \leq p \leq p+k \leq n$, corresponds to a sub-sequence ${\Lambda'}$ of ${\Lambda}$ (i.e., ${v_p\rightarrow v_{p+1}\rightarrow ...\rightarrow v_{p+k}}$). Moreover, an eligible span ${S'}$ also corresponds to a connected substructure ${T'}$ of ${T}$, which meet the following 2 requirements:
\begin{itemize}
    \item there is at most one node ${v_i\in \Lambda'}$ which is the child node of node ${v\in \Lambda \backslash \Lambda'}$\footnote{If there is no such node, we specifiy that the first node in the in-order traversal sequence is ${v_i}$.};
    \item there is at most one node ${v_o\in \Lambda'}$ which is the parent node of node ${v\in \Lambda \backslash \Lambda'}$;
\end{itemize}
Note that each node in the tree $T$ has one parent node and at least one child node. Specially, the parent node of the root node and the child node(s) of the leaf node(s) are special imaginary nodes.
\end{myDef}

Plus, we append the pseudo-code of SpanSub here in Algorithm~\ref{algo:spansub}. Note that: 

For SCAN task, we only substitute spans in the both the input side and target side simultaneously when there is no confusion:
\begin{itemize}
    \item If there are repetitively matched spans in either input side or output side, we substitute all of those repetitive ones at the same time. For example, input "walk and walk twice" is supposed to be interpreted as the target "W W W". If we are going to substitute "walk" with "jump" in the input side and its counterpart "W" with "J" in the target side, we are supposed to simultaneously substitute all of the matched spans, resulting in "jump and jump twice" $\rightarrow$ "J J J".
    \item If there are more than one kinds of span-matchs (in either input side or target side) and there is(are) overlap(s) between these matchs, we discard this example to alleviate the introduction of imprecise substitution. For example, input "walk around right thrice" is supposed to be interpreted as the target "<SOS> TR W TR W TR W TR W TR W TR W TR W TR W TR W TR W TR W TR W <EOS>" (supposing that we have already extracted the span "walk around right" $\rightarrow$ "TR W TR W TR W TR W"). However, we can not simultaneously substitute the "walk around right" in the input side and "TR W TR W TR W TR W" in the target side for there are many kinds of match (e.g., both of index[1, 5] and index[3, 7] are "TR W TR W TR W TR W".) in the target side and there exist overlaps between them.
\end{itemize}
Since GeoQuery is a highly realistic dataset (hence there are not always one-to-one mappings between words in the input sentences and words in the target programs, which potentially results generation of many noisy data.), we additionally impose two constraints to help with filter these generated noisy data: 1) if a modifier word in the target side(e.g., "largest\_one") could be mapped to several different words in the input side(e.g.,"largest", "most", ...), we need to pay attention when substituting the words(e.g., "area\_1") modified by this modifier or the modifier itself : we discard the synthetic new data covering the novel <modifier, modified word> combinations (e.g., "largest area" $\rightarrow$ "largest\_one ( area\_1 )", while "most area" makes no sense.); 2) if a modified word in the input side(e.g., "largest") could be mapped to several different words in the target side(e.g., "largest", "largest\_one" and "longest"), we can induce that words in the target side like "river" can only follow after "longest" if there is no case in the training set showing that "river" can follow after other interpretation of "largest" (i.e., "largest" and "largest\_one"). Hence we can directly discard those synthetic examples covering "largest ( river ( .." or "largest\_one ( river ( ..".

\begin{algorithm}[t]
\label{algo:spansub}
	\caption{\textbf{SpanSub}}
        \label{algo:spansub}
	\KwIn{Original dataset $\mathcal{D}$, the number of generated examples ${N}$, Span-Alignments extraction algorithm ${\mathcal{A}}$, Span-Classification function ${\Pi}$, Iterative Depth ${K}$.}
	\KwOut{Augmented dataset ${\mathcal{D}_{aug}}$.}  
	
        ${align}$, ${spans}$ $\leftarrow$ Run ${\mathcal{A}}$ on ${\mathcal{D}}$;  \\
        ${\mathcal{D}_{train}} \leftarrow\ \mathcal{D}$;\\
        \For{i $\leftarrow$ 1 to $K$}{
        ${\mathcal{D}_{aug}} \leftarrow\ \{\ \}$;\\
        
	\For{j $\leftarrow$ 1 to $N$}{
		   
	       Uniformly draw $d \in \mathcal{D}_{train}$ ;\\
        
        $({{inp},{out}})\leftarrow d $;\\
       
        Uniformly draw span ${s}$ from ${inp}$;\\
        
		Uniformly draw span ${s'}\in \{ v| v\in {spans}, {\Pi(v)} = {\Pi(s)}\}$;\\
  
            ${inp}_{aug} \leftarrow$ substitute ${s}$ with ${s'}$ in ${inp}$;\\
            
            ${out}_{aug} \leftarrow$ substitute ${align(s)}$ with ${align(s')}$ in ${out}$;\\
            
            $d_{aug}\leftarrow ({inp}_{aug},{out}_{aug}) $;\\
            
            ${\mathcal{D}_{aug}} \leftarrow {\mathcal{D}_{aug}} \cup \{d_{aug}\}$ \Comment{dedup}
	}
        $\mathcal{D}_{train} \leftarrow \mathcal{D}_{aug} \cup \mathcal{D}_{train}$;\\
        }
	\Return{${\mathcal{D}_{aug}}$}
\end{algorithm}

\subsection{L2S2 framework}
Here we also append the pseudo-code of training L2S2 framework in Algorithm~\ref{algo:l2s2}.
\label{append:l2s2_algorithm}
\begin{algorithm}[t]
	\caption{\textbf{Training L2S2 framework}}
        \label{algo:l2s2}
	\KwIn{Original dataset ${\mathcal{D}}$, \\L2S2 generator initialized parameters ${\phi_{0}}$, \\Seq-to-Seq Model initialized parameters ${\theta_{0}}$, \\Warm-up update number ${m}$,\\
    Sampled action number for each given example ${T}$. 
    }
	\KwOut{L2S2 generator parameters ${\phi_{f}}$, Seq-to-Seq Model parameters ${\theta_{f}}$.}  
	
        ${\theta \leftarrow \theta_{0}}$; ${\phi \leftarrow \phi_{0}}$
        
        \For{${step}\leftarrow$ 1 to m}{ 
            
            Sample ${\mathcal{B} \sim \mathcal{D}}$; 
            
            Optimize ${\theta}$ on ${\mathcal{B}}$ through Objective~\ref{obj:theta} 
        }
        
	\While{not converged}{
		  
            Sample ${\mathcal{B} \sim \mathcal{D}}$;
        
            \For{${t}\leftarrow$ 1 to T}{
            
            Sample ${\mathcal{B}_{gen,t} \sim p(\mathcal{B}_{gen}|\mathcal{B},\phi)}$;
            }
        
        Optimize ${\phi}$ on ${\{B_{gen,t}\}_{t=1}^T}$ through Objective~\ref{obj:phi} 
        
        Sample ${\mathcal{B} \sim \mathcal{D}}$;
        
        Sample ${\mathcal{B}_{gen} \sim p(\mathcal{B}_{gen}|\mathcal{B},\phi)}$;
        
        Optimize ${\theta}$ on ${B_{gen}}$ through Objective~\ref{obj:theta} 
	}
	\Return{${\phi,\theta}$}
\end{algorithm}

\section{Additional Experiments}
In this section, we mainly provide additional experiment results to support the conclusions in the main text(Section~\ref{append:experiments}). 
\label{append:experiments}
\subsection{The maximum numbers of distinct augmented examples with different augmentation methods on GeoQuery task}
As we discussed in Section~\ref{sec:intro}, we hypothesize that SpanSub enables multi-grained compositions of substantial substructures in the whole training set and thus lead to improvement for various kinds of compositional generalization.
\begin{table}
\centering
\resizebox{0.47\textwidth}{!}{
\begin{tabular}{c|c|c|c|c}
\hline
\footnotesize{w/o Aug} & \footnotesize{GECA} &  \footnotesize{LexSym} & \footnotesize{SUBS}& \footnotesize{SpanSub} \\
\hline
$519$ & $2,028$ & $28,520$ & $20,564$ & $99,604$ \\
\hline 
\end{tabular}
}
\caption{
The maximum numbers of distinct augmented examples on the query split of GeoQuery dataset with different augmentation methods. w/o Aug refers to the number of original training examples.
}
\label{tab:max_aug_num}
\end{table}
We provide a statistic on the maximum number of augmented examples (after deduplication) on the query split of GeoQuery dataset with different augmentation methods, including GECA, LexSym, SUBS and SpanSub in Table~\ref{tab:max_aug_num}. SpanSub overwhelmingly outweigh other augmentation methods and even their summation, which reflects its superiority of exploring potential compositions of substantial substructures in the whole training set.

\begin{table}
\centering
\resizebox{0.47\textwidth}{!}{
\begin{tabular}{lcccc}
\hline
Error Type & walk right & walk opposite right & walk around right\\
\hline
RandS2 & $51.2\%$&$28.1\%$ & $76.8\%$ \\
L2S2 & $37.4\%$ & $14.6\%$ & $40.2\%$\\
\hline 
\end{tabular}
}
\caption{
Comparision of the error rates($\downarrow$) of examples with different concepts (i.e., spans) between RandS2 and L2S2. Results are attained using the same LSTM architecture with ~\cite{geca} on SCAN-MCD3 split.
}
\label{tab:w_a_r_err_dec}
\end{table}
\subsection{Ablation Studies and Control Experiments}
~\label{appendix:ablation}
In this section, we investigate the effect of SpanSub, L2S2 framework training and their combination. Besides, we also investigate the effectiveness of the optimizable L2S2 augmentor in the L2S2 framework through control experiments. Our results are shown in Table~\ref{tab:ablation_exps}.
\paragraph{Effectiveness of SpanSub and L2S2 framework training}
Through observing the experiment results of "LSTM"-group, "+L2S2"-group, "+SpanSub"-group and "+SpanSub+L2S2"-group on SCAN MCD(1,2,3) and COGS tasks, we can induce a consistent conclusion that : (1) both of the SpanSub data augmentation method and the L2S2 framework training method can improve the performance of our base model and (2) the combination of them, SpanSub+L2S2, can further boost the performance of our base model. These empirically verify the effectiveness of both SpanSub and L2S2 parts. 
\paragraph{Effectiveness of L2S2 augmentor in L2S2 framework}
Furthermore, to verify the the effectiveness of the optimizable L2S2 augmentor part in the L2S2 framework, we design control experiments where the L2S2 augmentor is substituted with a non-differentiable random augmentor (The function of random augmentor is to randomly substitute a span in the given example with another span in the span set.) and everything else is maintained (We name it "RandS2"). Through observing the results of "+SpanSub", "+SpanSub+RandS2" and "+SpanSub+L2S2", we can draw a conclusion that RandS2 is not capable of functioning as L2S2 when being combined with SpanSub and in some cases RandS2 even has slight negative influence on SpanSub. 
Through observing the results of "+RandS2" and "+L2S2", we can similarly induce that RandS2 can not work as well as L2S2 on SCAN-MCD splits when being utilized alone . The reason for RandS2 can also improve the performance of based models is that RandS2 can be viewed as an online version SpanSub here.
To conclude, we empirically verify the effectiveness of L2S2 augmentor in L2S2 framework through comparing the effect of it with the effect of a random augmentor.
\subsection{Experiments with different kinds of Base Models}
\begin{table}
\centering
\resizebox{0.47\textwidth}{!}{
\begin{tabular}{lcccc}
\hline
Method & MCD1 & MCD2 & MCD3 \\
\hline
\hline
\textbf{\emph{LSTM}$_1$} &8.9\%\footnotesize{$\pm$ 1.6\%} &11.9\%\footnotesize{$\pm$ 9.4\%}  & 6.0\%\footnotesize{$\pm$ 0.9\%}  \\
\hline
+RandS2 &46.6\%\footnotesize{$\pm$ 8.9\%} &52.3\%\footnotesize{$\pm$ 2.4\%}  & 58.8\%\footnotesize{$\pm$ 3.1\%} \\
+L2S2 & \textbf{55.1\%}\footnotesize{$\pm$ 17.6\%} & \textbf{54.3\%}\footnotesize{$\pm$ 8.0\%}  & \textbf{70.8\%}\footnotesize{$\pm$ 5.0\%}\\
\hline
+SpanSub &63.4\%\footnotesize{$\pm$ 13.1\%} &72.9\%\footnotesize{$\pm$ 10.1\%}  & 74.0\%\footnotesize{$\pm$ 10.2\%}\\
+SpanSub+RandS2 &63.3\%\footnotesize{$\pm$ 11.7\%} &66.2\%\footnotesize{$\pm$ 6.6\%}  & 71.2\%\footnotesize{$\pm$ 13.9\%} \\
+SpanSub+L2S2 &\textbf{67.4\%}\footnotesize{$\pm$ 12.1\%} &\textbf{73.0\%}\footnotesize{$\pm$ 10.1\%}  & \textbf{80.2\%}\footnotesize{$\pm$ 1.8\%} \\
\hline
\hline
\textbf{\emph{LSTM}$_2$}  &6.8\%\footnotesize{$\pm$ 3.5\%} &9.6\%\footnotesize{$\pm$ 3.0\%}  & 9.3\%\footnotesize{$\pm$ 2.5\%}  \\
\hline
+RandS2 &41.4\%\footnotesize{$\pm$ 4.2\%} & 64.1\%\footnotesize{$\pm$ 7.6\%}  & 70.1\%\footnotesize{$\pm$ 5.4\%} \\
+L2S2 & \textbf{44.3\%}\footnotesize{$\pm$ 6.7\%} & \textbf{65.9\%}\footnotesize{$\pm$ 6.7\%}  & \textbf{76.5\%}\footnotesize{$\pm$ 4.3\%}\\
\hline
+SpanSub &52.7\%\footnotesize{$\pm$ 5.1\%} & 71.0\%\footnotesize{$\pm$ 6.4\%}  & 78.9\%\footnotesize{$\pm$ 2.3\%}\\
+SpanSub+RandS2 &55.1\%\footnotesize{$\pm$ 6.4\%} & 73.4\%\footnotesize{$\pm$ 6.5\%}  & 78.5\%\footnotesize{$\pm$ 6.2\%}\\
+SpanSub+L2S2 & \textbf{55.4\%}\footnotesize{$\pm$ 8.6\%} & \textbf{74.1\%}\footnotesize{$\pm$ 5.5\%}  & \textbf{80.8\%}\footnotesize{$\pm$ 7.4\%} \\
\hline
\hline
\emph{\textbf{Transformer}}  &1.7\%\footnotesize{$\pm$ 0.7\%} & 4.3\%\footnotesize{$\pm$ 1.3\%}  & 4.4\%\footnotesize{$\pm$ 1.2\%}  \\
\hline
+RandS2 &11.2\%\footnotesize{$\pm$ 2.2\%} & 37.0\%\footnotesize{$\pm$ 7.1\%}  & 48.1\%\footnotesize{$\pm$ 2.6\%}\\
+L2S2 &\textbf{19.3\%}\footnotesize{$\pm$ 2.2\%} & \textbf{68.1\%}\footnotesize{$\pm$ 1.7\%}  & \textbf{57.8\%}\footnotesize{$\pm$ 2.2\%}\\
\hline
+SpanSub &24.8\%\footnotesize{$\pm$ 1.7\%} & 79.4\%\footnotesize{$\pm$ 1.5\%}  & 61.3\%\footnotesize{$\pm$ 0.9\%}\\
+SpanSub+RandS2 &21.0\%\footnotesize{$\pm$ 1.9\%} & \textbf{80.2\%}\footnotesize{$\pm$ 2.3\%}  & 60.3\%\footnotesize{$\pm$ 1.3\%}\\
+SpanSub+L2S2 &\textbf{27.0\%}\footnotesize{$\pm$ 4.4\%} & \textbf{80.2\%}\footnotesize{$\pm$ 1.9\%}  & \textbf{63.3\%}\footnotesize{$\pm$ 2.3\%}\\
\hline 
\end{tabular}
}
\caption{
Experiments on SCAN-MCDs splits with three standard seq-to-seq models with different architectures.
Note that \textbf{\emph{LSTM}$_1$} is the LSTM-based seq-to-seq model in align with~\cite{geca} (base on one-layer LSTM and embedding dimension of 64) and \textbf{\emph{LSTM}$_2$} is the LSTM-based seq-to-seq model in align with~\cite{lexsym} (based on two-layer LSTM and embedding dimension of 512). \textbf{\emph{Transformer}} is the standard seq-to-seq model introduced by~\cite{transformer}. Here we use a transformer adopted from ~\cite{mutual}, with a three-layer encoder and a three-layer decoder (both encoder layers and decoder layers contain self-attention layers and fully-connected layers).
}
\label{tab:different_archs_exps}
\end{table}
A significant advantage of our SpanSub and L2S2 is their model-agnostic~\footnote{Here the term of model means the down-steam sequence-to-sequence model.} property so that we can easily apply these techniques to various base models with different architectures. 
In this section, we aim to answer the question that whether our proposed SpanSub and L2S2 methods can  consistently help improve the compositional generalization of standard base models with different architectures(e.g., LSTM seq-to-seq models with different architectures, and Transformer~\cite{transformer}) or not?\par
Firstly, we have empirically demonstrated the effectiveness of both proposed SpanSub and L2S2 methods on SCAN (standard splits and MCD splits) tasks with LSTM-based seq-to-seq model (in line with ~\cite{geca})and COGS task with another distinct LSTM architecture ( in line with ~\cite{lexsym}) respectively in Section~\ref{sec:main_result}. Moreover, here we conduct more experiments on SCAN-MCD splits with LSTM architecture (in line with ~\cite{lexsym}) and Transformer to demonstrate that Span and L2S2 can consistently help improve the compositional generalization of standard base models with different architectures.
Our results are shown in Table~\ref{tab:different_archs_exps}. Through observing these results, we find that our previous conclusions consistently hold with these three different standard seq-to-seq models (i.e., \textbf{\emph{LSTM$_1$}}, \textbf{\emph{LSTM$_2$}} and \textbf{\emph{Transformer}}), which stands for that both SpanSub and L2S2 can help various down-stream sequence models better compositionally generalize.

\begin{table*}[h!]
\centering
\resizebox{\linewidth}{!}{
\begin{tabular}{lccc||c}
\hline
Method & MCD1 & MCD2 & MCD3 & COGS\\
\hline
LSTM  &8.9\%\footnotesize{$\pm$ 1.6\%} &11.9\%\footnotesize{$\pm$ 9.4\%}  & 6.0\%\footnotesize{$\pm$ 0.9\%} & 55.4\%\footnotesize{$\pm$ 4.2\%} \\
\hline
+RandS2 \footnotesize{(\textbf{Control})}  &46.6\%\footnotesize{$\pm$ 8.9\%} &52.3\%\footnotesize{$\pm$ 2.4\%}  & 58.8\%\footnotesize{$\pm$ 3.1\%} & \textbf{89.7}\%\footnotesize{$\pm$ 0.2\%}  \\
+L2S2 \footnotesize{(\textbf{Ours})}  &\textbf{55.1}\%\footnotesize{$\pm$ 17.6\%} &\textbf{54.3}\%\footnotesize{$\pm$ 8.0\%}  & \textbf{70.8}\%\footnotesize{$\pm$ 5.0\%} & \textbf{89.7}\%\footnotesize{$\pm$ 0.2\%}  \\
\hline
+SpanSub \footnotesize{(\textbf{Ours})}  &63.4\%\footnotesize{$\pm$ 13.1\%} &72.9\%\footnotesize{$\pm$ 10.1\%}  & 74.0\%\footnotesize{$\pm$ 10.2\%}  & 91.8\%\footnotesize{$\pm$ 0.1\%} \\
+SpanSub+RandS2\footnotesize{(\textbf{Control})}  &63.3\%\footnotesize{$\pm$ 11.7\%} &66.2\%\footnotesize{$\pm$ 6.6\%}  & 71.2\%\footnotesize{$\pm$ 13.9\%} & 91.9\%\footnotesize{$\pm$ 0.1\%} \\
+SpanSub+L2S2 \footnotesize{(\textbf{Ours})}  &\textbf{67.4\%}\footnotesize{$\pm$ 12.1\%} &\textbf{73.0\%}\footnotesize{$\pm$ 10.1\%}  & \textbf{80.2\%}\footnotesize{$\pm$ 1.8\%} & \textbf{92.3}\%\footnotesize{$\pm$ 0.2\%} \\
\hline 
\end{tabular}
}
\caption{
Ablation studies of SpanSub and L2S2 and comparison with control group(RandS2).
}
\label{tab:ablation_exps}
\end{table*}

\end{document}